\documentclass[10pt,twocolumn,letterpaper]{article}

\usepackage[pagenumbers]{cvpr} %

\usepackage{graphicx}
\usepackage{amsmath}
\usepackage{amssymb}
\usepackage{booktabs}
\usepackage{pifont}%
\newcommand{\cmark}{\ding{51}}%
\newcommand{\xmark}{\ding{55}}%
\usepackage{mathabx}
\usepackage{enumitem}
\usepackage{soul}

\newcommand{\set}[1]{\{{#1}\} }
\usepackage[utf8]{inputenc}

\usepackage[pagebackref,breaklinks,colorlinks]{hyperref}
\usepackage[dvipsnames]{xcolor}
\usepackage[capitalize]{cleveref}
\crefname{section}{Sec.}{Secs.}
\Crefname{section}{Section}{Sections}
\Crefname{table}{Table}{Tables}
\crefname{table}{Tab.}{Tabs.}

\def\cvprPaperID{5885} %
\def\confName{CVPR}
\def\confYear{2023}

\begin{document}

\title{
Teaching CLIP to Count to Ten
}

\author{
\begin{tabular}{cccc}
    Roni Paiss$^{1, 2}$ \quad
    Ariel Ephrat$^{1}$ \quad
    Omer Tov$^{1}$ \quad
    Shiran Zada$^{1}$ \\
    Inbar Mosseri$^{1}$ \quad
     Michal Irani$^{1, 3}$\quad
    Tali Dekel$^{1, 3}$
\end{tabular} \\
\small{$^1$Google Research \qquad \qquad $^2$Tel Aviv University \qquad \qquad $^3$Weizmann Institute of Science}\vspace{-1.3cm}
}
\twocolumn[{%
\renewcommand\twocolumn[2][]{#1}%
\maketitle%

\vspace*{-0.3cm}
\begin{center}
    \centering
    \captionsetup{type=figure}
    \includegraphics[width=0.98\textwidth]{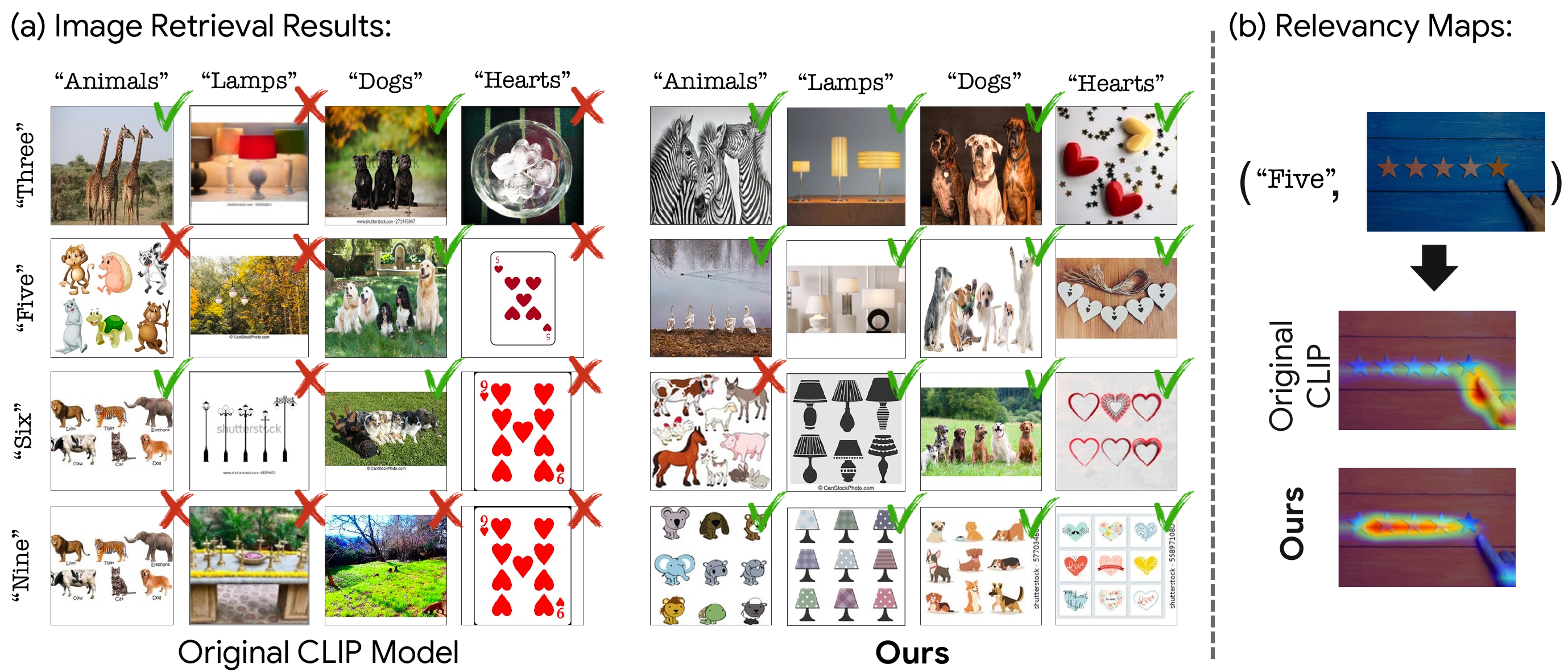}
    \vspace*{-0.2cm}\captionof{figure}{
    {\bf Counting-aware CLIP.} \it 
    We demonstrate the effectiveness of our improved CLIP by showing: (a)~image retrieval using text captions with different types of objects and their counts in the image (images that match the caption are marked with {\color{OliveGreen}\cmark} and images that do not match it are marked with {\color{Mahogany}\xmark}).  Our model retrieves images that match the requested number of objects, while the baseline CLIP often retrieves images that depict the wrong number of objects, or images where the number is explicitly written in the image (e.g. ''nine hearts" - the image contains the number "9", but has 11 hearts). (b)~Attention maps demonstrating that our model focuses its attention to all matching object instances in the image, as opposed to the original CLIP.
    }\label{fig:teaser_fig}
\end{center}%
\vspace*{-0.2cm}
}] 

\newcommand\blfootnote[1]{%
  \begingroup
  \renewcommand\thefootnote{}\footnote{#1}%
  \addtocounter{footnote}{-1}%
  \endgroup
}

\blfootnote{The first author performed this work as an intern at Google Research.}
\blfootnote{Project page: \url{https://teaching-clip-to-count.github.io/}}

\begin{abstract}
\vspace{-0.3cm}
Large vision-language models (VLMs), such as CLIP, learn rich joint image-text representations, facilitating advances in numerous downstream tasks, including zero-shot classification and text-to-image generation. Nevertheless, existing VLMs exhibit a prominent well-documented limitation -- they fail to encapsulate compositional concepts such as counting.  
 We introduce a simple yet effective method to improve the quantitative understanding of VLMs, while maintaining their overall performance on common benchmarks. Specifically, we propose a new counting-contrastive loss used to finetune a pre-trained VLM in tandem with its original objective. Our counting loss is deployed over automatically-created counterfactual examples, each consisting of an image and a caption containing an incorrect object count.  For example, an image depicting three dogs is paired with the caption ``Six dogs playing in the yard''. Our loss encourages discrimination between the correct caption and its counterfactual variant which serves as a hard negative example. To the best of our knowledge, this work is the first to extend CLIP's capabilities to object counting.
Furthermore, we introduce ``CountBench'' -- a new image-text counting benchmark for evaluating a model's understanding of object counting. We demonstrate a significant improvement over state-of-the-art baseline models on this task. Finally, we leverage our count-aware CLIP model for image retrieval and text-conditioned image generation, demonstrating that our model can produce specific counts of objects more reliably than existing ones.
\end{abstract}

\section{Introduction}

Since the advent of CLIP~\cite{clip}, training large vision-language models (VLMs) has become a prominent paradigm for representation learning in computer vision. By observing huge corpora of paired images and captions crawled from the Web, these models learn a powerful and rich joint image-text embedding space, which have been employed in numerous visual tasks, including classification~\cite{zhou2022coop, zhou2022cocoop}, segmentation~\cite{Zabari2021SemanticSI, Li2022LanguagedrivenSS}, motion generation~\cite{tevet2022motionclip}, image captioning~\cite{tewel2021zero, mokady2021clipcap}, text-to-image generation~\cite{Nichol2022GLIDETP, vqgan+clip, Liu2021FuseDreamTT, Saharia2022PhotorealisticTD, Ramesh2021ZeroShotTG} and image or video editing~\cite{Vinker2022CLIPassoSO, Gal2022StyleGANNADA, patashnik2021styleclip, Avrahami2022BlendedDF, Chefer2022ImageBasedCE, BarTal2022Text2LIVETL, kawar2022imagic}. 
Recently, VLMs have also been a key component in text-to-image generative models~\cite{imagen, Ramesh2021ZeroShotTG, balaji2022eDiff-I, ramesh2022hierarchical}, which rely on their textual representations to encapsulate the rich and semantic meaning of the input text prompt.

Despite their power, prominent VLMs, such as  CLIP~\cite{clip} and BASIC~\cite{BASIC}, are known to possess a weak understanding of compositional concepts, such as the relation between objects or their number present in the image~\cite{clip, thrush_and_ross2022winoground,Liu2021LearningTC}. This is demonstrated in \cref{fig:teaser_fig}, where, when given a caption of the template ``a photo of $\{number\}$ $\{objects\}$'', CLIP often fails to retrieve images that correctly match the described number. Downstream applications that rely on VLM-based representations inherit these limitations, e.g., image generation models struggle to reliably produce specific counts of objects~\cite{imagen_blog}. 

In this work, we focus on the counting task  and introduce a novel method that enhances the quantitative understanding of large-scale VLMs by encouraging them to produce representations that are sensitive to the number of objects in the image and text.

We hypothesize that the reason existing VLMs fail to learn the concept of counting is twofold: ($i$) 
Captions that accurately specify the number of objects become extremely rare in the data as the number of objects increases. For example, we found that for more than six objects, captions would typically contain a general form of quantity, e.g., ``a group of ...'' or ``many ..'', rather than an accurate count. ($ii$)~Even with such examples in hand, the task of counting, i.e., associating the visible number of objects in an image with the number in the caption, does not sufficiently contribute to the VLM's discriminative training objective. This is because other textual and visual features (e.g., nouns and object categories) are more informative for associating an image with its true caption. 

We thus suggest to mitigate each of these problems by: ($i$) Creating suitable training data in which the captions contain accurate numbers of objects. ($ii$) Designing a training objective whereby understanding object counts is critical for discriminating between the correctly associated caption and incorrect ones. 

\begin{figure*}
    \centering
    \includegraphics[width=0.90\linewidth]{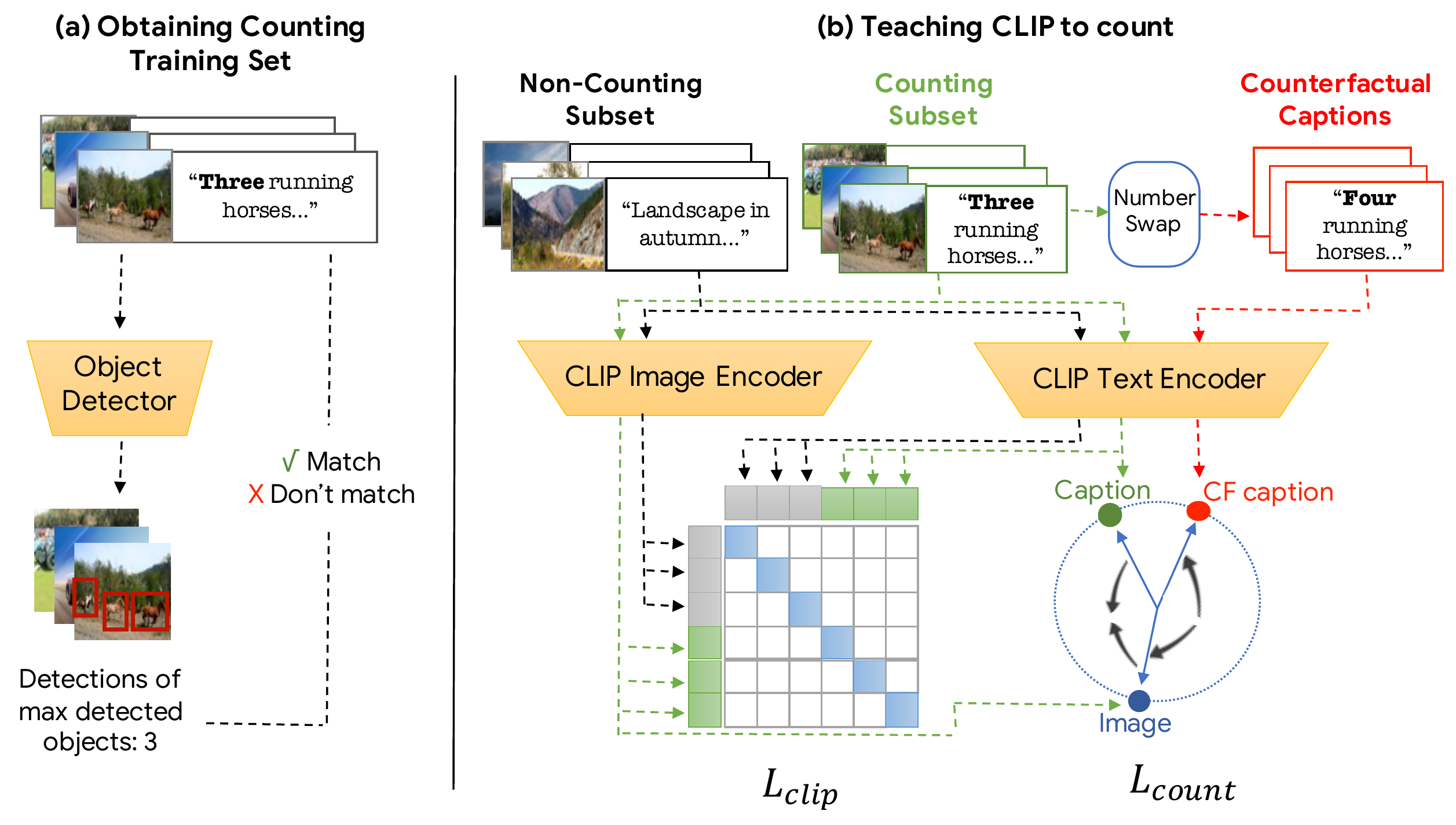}
  \vspace*{-0.1cm}
    \caption{{\bf Method overview} (a) We create a text-image counting training set in which each caption expresses the number of objects depicted in the corresponding image.  This is done by using an off-the-shelf object detector to automatically identify text-image examples in which the text count matches the number of visible objects in the image (see Sec.~\ref{sec:method-data}). (b) We finetune a pre-trained CLIP model using our counting subset (a),  through a dedicated  contrastive objective $L_{count}$, used in addition to the original (general) text-image contrastive objective ($L_{clip}$). Specifically, given a text-image example from our counting subset, we automatically create a counterfactual prompt by replacing the true object count in the original caption with an incorrect count; $L_{count}$ encourages the model to embed the image close to its original caption embedding (expressing the true object count) and far from its counterfactual count. (see Sec.~\ref{sec:method-loss}).}
    \label{fig:arch}
\end{figure*}

More specifically, as illustrated in Fig. \ref{fig:arch}, we automatically create a clean and diverse \emph{counting training set} by curating image-text examples where the image depicts multiple objects and its caption expresses their count (e.g., Fig.~\ref{fig:countbench}). To do so, we employ off-the-shelf computer vision tools to cross-validate the number of observed objects in the image with the textual number in the caption. We then finetune a pretrained VLM by formulating counting as  a discriminative task -- for each example, we create a counterfactual caption by swapping the spelled number associated with the object count with a different randomly selected number. The model's objective is then to associate the image correctly with its true count caption, discriminating it from the negative one.

\begin{figure*}
    \centering
    \includegraphics[width=0.85\textwidth]{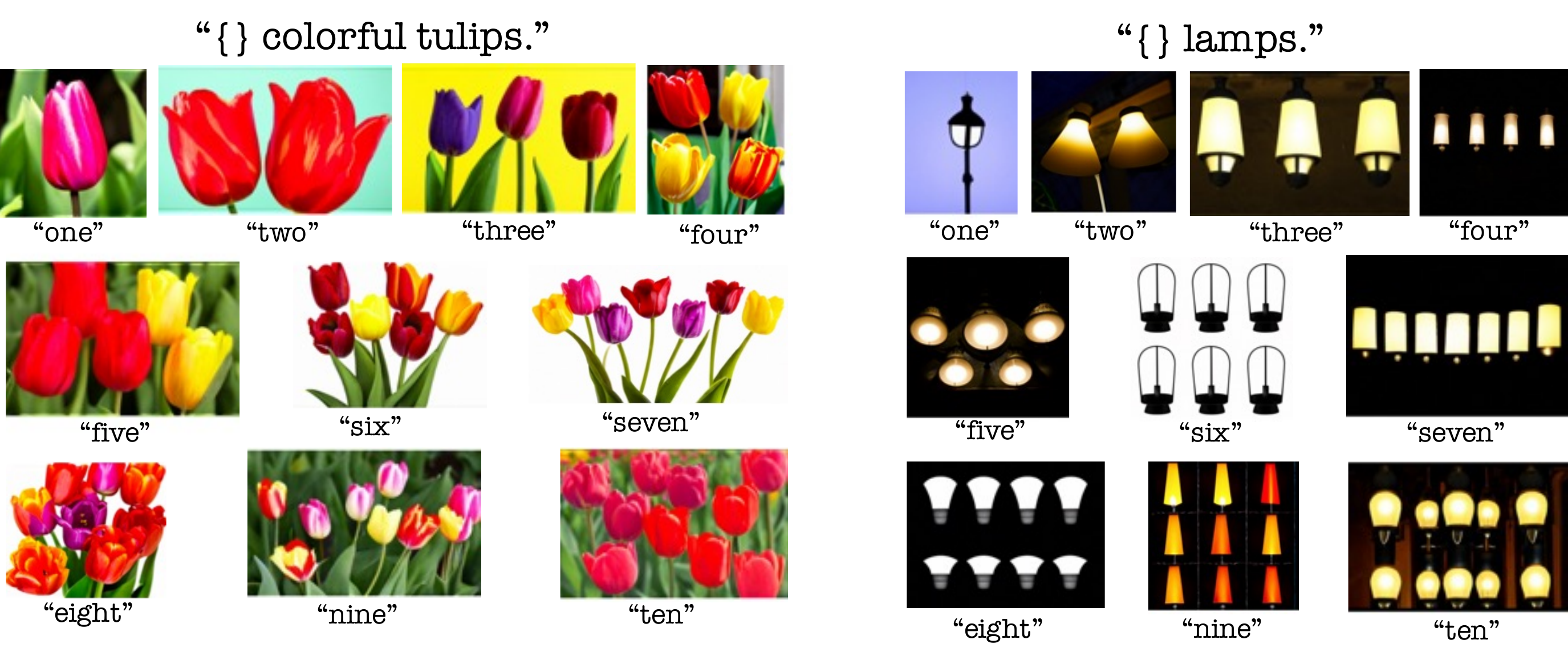}
  \vspace*{-0.3cm}
    \caption{\textbf{Generated samples with Imagen using our counting-aware CLIP as backbone}. \it The Imagen model benefits from the counting-aware representations produced by the our CLIP model, and is able to generate images that accurately follow the amounts specifies in the captions}
    \label{fig:text-to-image}
  \vspace*{-0.3cm}
\end{figure*}

To evaluate our method, we introduce {\it CountBench} -- a carefully curated object counting benchmark, consisting of 540 diverse, high quality image-text examples.
We evaluate our method on two prominent contrastive VLMs: CLIP~\cite{clip} and BASIC~\cite{BASIC}, and demonstrate a significant improvement in accuracy in the task of zero-shot count classification over baseline models.
Importantly, we achieve this while maintaining the original knowledge learned by the VLM, as demonstrated by an extensive evaluation of our model on standard zero-shot downstream tasks. The quantitative understanding of our model is further evident by our text-to-image retrieval results (e.g., Fig.~\ref{fig:teaser_fig}(a)), as well as by the relevancy maps of our model, which demonstrate that the model correctly attends to all visible objects whose count is specified in the text (e.g., Fig.~\ref{fig:teaser_fig}(b)).  Finally, we train a large-scale text-to-image generative model~\cite{imagen} which incorporates our counting training set and finetuned CLIP text encoder.  The generated images from this model exhibit higher fidelity to the  number of objects specified in the input prompts (Fig.~\ref{fig:text-to-image}).

\noindent To summarize, our main contributions are:
\begin{enumerate}
[topsep=1ex,itemsep=0.5ex,partopsep=1ex,parsep=1ex,leftmargin=0.5cm]
    \item  A novel training framework for tackling the task of vision-language counting -- an important limitation of current VLMs.
    \item  A new benchmark, ``\emph{CountBench}'', carefully filtered and validated for evaluating VLMs on the counting task.
    \item We apply our method to the widely-adopted VLMs, CLIP~\cite{clip} and BASIC~\cite{BASIC}, demonstrating significant improvement on the counting task, while maintaining zero-shot accuracy on common benchmarks.
    \item We utilize our counting-aware VLMs for downstream tasks including  image retrieval and text-to-image generation, demonstrating more reliable results when the text prompt contains a specific number of objects.
\end{enumerate}

\section{Related work}
\paragraph{Contrastive vision-language models}
Vision-language models have demonstrated impressive success in vision and multimodal tasks~\cite{clip,BASIC,Alayrac2022FlamingoAV,Chen2022PaLIAJ,Singh2021FLAVAAF}. These models are trained on huge image-text datasets, and applied for downstream applications in a zero-shot manner or via finetuning.
In this work, we focus on contrastive VLMs, such as CLIP~\cite{clip} and BASIC~\cite{BASIC}, as they are widely used both for downstream applications and as backbones for generative vision-language models~\cite{imagen,Ramesh2022HierarchicalTI}.
CLIP~\cite{clip} is trained on 400~million pairs of images and captions collected from the Web, using a contrastive objective, where matching text-image pairs should have a low cosine distance, and non-matching texts and images should be far apart.
The model consists of a transformer~\cite{vaswani2017attention} text backbone and  a ViT~\cite{dosovitskiy2020image} or ResNet~\cite{He2016DeepRL} vision backbone. 
The representations computed by CLIP have proven to be very effective in vision and multimodal tasks, due to their zero-shot capabilities and semantic nature, and have been widely used as a prominent component in numerous tasks and methods.
BASIC~\cite{BASIC} scaled up the size of the model, batch size and dataset, improving zero-shot accuracy on common benchmarks, and uses CoAtNet~\cite{Dai2021CoAtNetMC} for its vision backbone.

\vspace*{-0.3cm}
\paragraph{Compositionality and counting in vision-language models}
While demonstrating impressive recognition capabilities, large VLMs such as CLIP\cite{clip} and BASIC~\cite{BASIC} are known to only partially capture the meaning of the text. Numerous works~\cite{clip, thrush_and_ross2022winoground, Liu2021LearningTC} have shown that they fail to understand compositional concepts, such as the relation between objects or their number in the image.  Paiss et al.~\cite{Paiss2022NoTL} demonstrated that CLIP attends to only a small subset of its input, mainly the nouns, and often ignores adjectives, numbers and prepositions. 

Counting has remained a stand-alone task under the domain of visual question answering (VQA), tackled with specifically designed architectures and techniques.
Some approaches used are counting-specific architectures, such as a layer that infers the number of objects from the normalized attention weights \cite{Zhang2018LearningTC}, relation networks to model the relations between foreground and background regions \cite{Acharya2018TallyQAAC}, and others \cite{Nguyen2021MOVIERM}. Our work defers from these prior efforts in several key aspects: ($i$) While previous efforts are restricted to VQA architectures and problem formulation, our goal is to improve the quantitative understanding of general-purpose contrastive VLMs (e.g., CLIP and BASIC), used in various vision and multimodal tasks where counting-aware solutions are not currently available. ($ii$) Our work can enhance the zero-shot counting capabilities of VLMs to unrestricted objects, unlike prior methods that are trained on specific domains, which can be problematic for new domains where no counting labels are available.

\vspace*{-0.3cm}
\paragraph{Text-conditioned generation}
The field of text-to-image generation has made significant progress in recent years, mainly using CLIP as a representation extractor. Many works use CLIP to optimize a latent vector in the representation space of a pretrained GAN \cite{vqgan+clip, Liu2021FuseDreamTT, Gal2022StyleGANNADA, patashnik2021styleclip}, others utilize CLIP to provide classifier guidance for a pretrained diffusion model \cite{Avrahami2022BlendedDF}, and \cite{BarTal2022Text2LIVETL} employ CLIP to optimize a Deep Image Prior model~\cite{ulyanov2018deep} that correctly edits an image. Recently, the field has shifted from employing CLIP as a loss network for optimization, and into using it as a backbone in huge generative models~\cite{imagen, Ramesh2022HierarchicalTI}, resulting in impressive photorealistic results.
However, these methods inherit the limitations of the VLMs. Text-to-image generation methods that use CLIP fail to reliably produce specific counts of objects
   ~\cite{imagen, Yu2022ScalingAM}, and understand syntactic processes~\cite{Leivada2022DALLE2F, Rassin2022DALLE2IS}. 
While several attempts have been made to improve the correspondence of text-guided generated images~\cite{materzynska2022disentangling, Du2020CompositionalVG}, they focus on the generative pipeline, while we attempt to improve the text representations themselves.

\section{Method}
\label{sec:method}
Our goal is to teach a pre-trained VLM (e.g., CLIP) to count, i.e., to improve its quantitative textual and visual understanding. Our framework, illustrated in \cref{fig:arch}, consists of two main stages. We first automatically create a \emph{counting training set}, comprising clean and diverse images along with corresponding captions that describe the number of visible objects in the scene. We then leverage this dataset to finetune the VLM through a designated count-based contrastive loss that is used in tandem with the original generic image-text objective.

More specifically, our key idea is to automatically generate counterfactual examples by swapping the true object count in the caption with a different random number. Our new counting loss encourages the model to embed an image close to its true count, as expressed by the original caption, while pushing it away from the embedding of the counterfactual count prompt. As the only difference between the correct caption and their counterfactual counterparts is a single word---the spelled number of objects---the model has to distinguish between  the correct and incorrect count in order to succeed in its training task.
Next, we describe our dataset creation and finetuning paradigm in detail.

\subsection{Creating an image-text counting train set}
\label{sec:method-data}
\begin{figure}
    \centering
    \includegraphics[width=\linewidth]{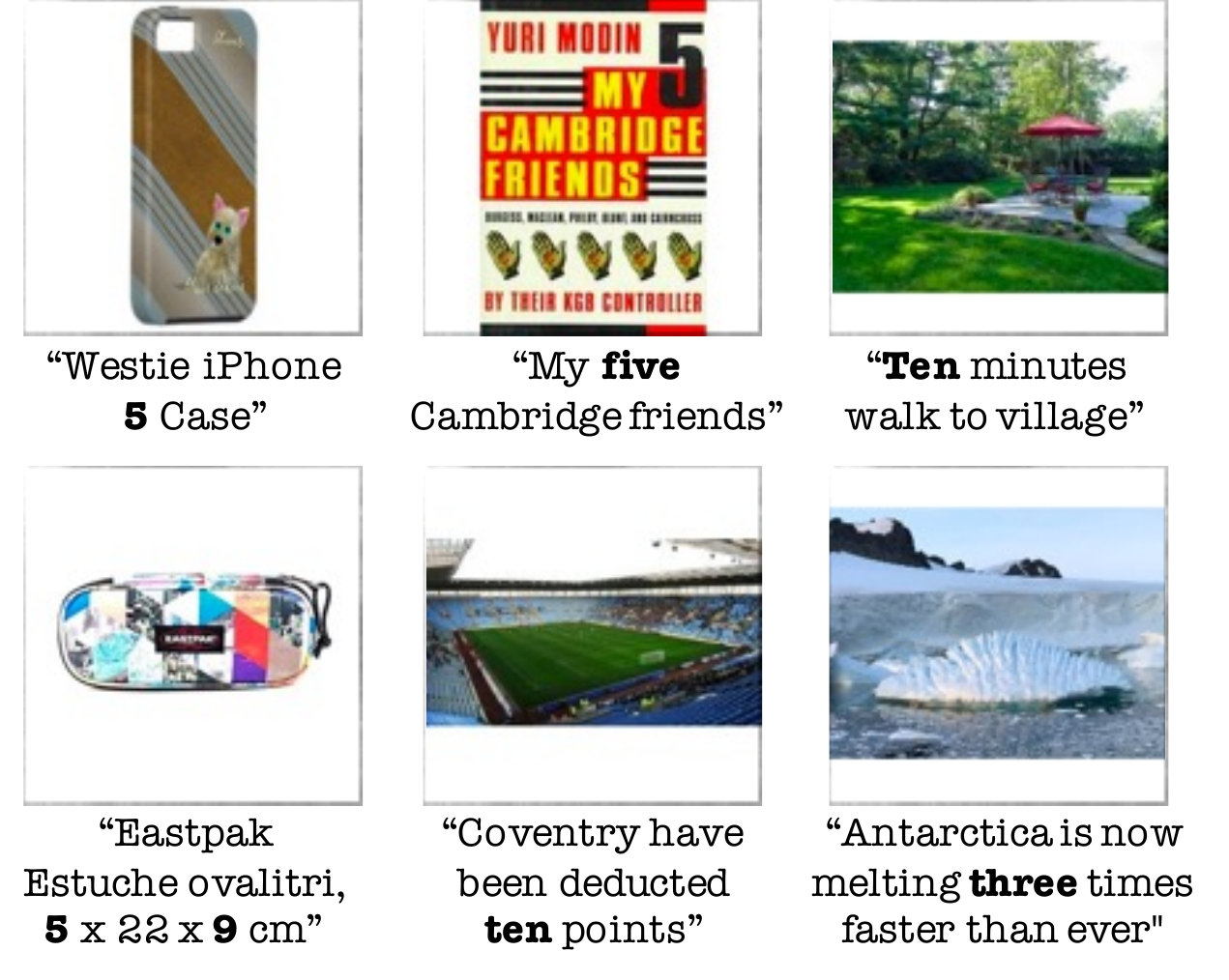}
  \vspace*{-0.5cm}
    \caption{{\bf Examples of image captions where the numbers are NOT related to object counts. These are automatically \mbox{filtered-out} by our method.} In all above  examples the numbers indicated in the caption do not refer to an actual object count. Numbers often specify measures, versions, dates, time, written numbers in the image, or numbers that refer to things not visible in the image.}
    \label{fig:bad-nums}
  \vspace*{-0.3cm}
\end{figure}

A na\"ive approach for obtaining an image-text counting dataset is to filter a large-scale dataset by considering only the examples in which the caption contains a number. However, this approach results in a highly noisy dataset, since the number in the caption often refers to other attributes that are unrelated to counting, such as age, time, addresses etc, as seen in \cref{fig:bad-nums}. 

Recall that the crux of our method is a constrastive loss w.r.t. hard negatives which differ from the original caption only by the number of objects described. Thus, it is critical to ensure that a given image-text pair not only contains a number, but also that the number correctly refers to the number of instances of a particular object in the image. To verify these conditions, we employ several stages of automatic filtering in our data pipeline (\cref{fig:arch}~(a)):

First, we filter out all examples whose caption does not contain a spelled number $\in\set{``two", \dots,``ten"}$.
We do so, as we observed that non-spelled numbers, or numbers higher than ten, mostly appear in conjunction with a measure of time, (e.g. dates) or addresses, rather than numbers of objects present in the image. 

In the second stage, we verify that the spelled numbers indeed serve as object counters, and that the counted objects are visible and detectable in the image.
For example, for the caption ``A photo of \textit{three} dogs'', we verify that the image indeed depicts three visible dogs, no more, and no less. Only then can we use this as a \underline{positive caption}, and replace the number to create \underline{negative captions}, e.g., ``A photo of {\it five} dogs''. This count verification is achieved automatically by first applying an off-the-shelf object detector \cite{howard2019searching}, and counting the number of detections per object. We assume that the caption refers to the most prevalent object in the image. Thus, we retain only examples for which the number specified in the caption aligns with the number of instances of the maximally-detected object.  
We denote by $C$ our automatically filtered train set.

Naturally, the filtered data $C$ is unbalanced. 
The number of examples that pass our filtering drops significantly as the count increases, e.g., the number of  $``ten"$ image-text pairs is around $1000\times$ smaller than $``two"$. Training with such imbalanced data creates a bias---the loss can be reduced by classifying frequent numbers as the correct caption and rare numbers as counterfactual, regardless of the image content. 
Therefore, balancing the data is of essence.
Due to scarcity of examples depicting more than six objects, we choose to balance the numbers $``two"-``six"$ separately from the higher numbers  $``seven"-``ten"$. For each of the numbers $``two"-``six"$, we sample around $37K$ samples, while for $``seven"-``ten"$, we use all the samples passed by our filter. There are approximately $7K$ samples for $``seven"$ down to around $1.5K$ samples for $``ten"$. We found this approach to provide us with a diverse and relatively balanced training dataset, yet more sophisticated methods could be considered in the future. From this point on, $C$ will denote our filtered and balanced numbered training set.

\subsection{Teaching CLIP to count}
\label{sec:method-loss}
Our goal is to improve the quantitative understanding of a pre-trained VLM (e.g., CLIP), while preserving its real-world knowledge, as reflected by its zero-shot capabilities on commonly-evaluated benchmark tasks. Therefore, 
we use a combination of two loss functions: 
\begin{equation}
      L = L_{CLIP} + \lambda L_{count}
\end{equation}
where $L_{CLIP}$ is the regular contrastive loss of CLIP, $L_{count}$ is our counting-designated loss (described below), and $\lambda$ is a hyperparameter used to weight the two losses.

We finetune the model on two training sets: ($i$) A very large dataset collected from the Web that contains general in-the-wild images and captions. ($ii$) Our filtered numbered training set $C$, described in~\cref{sec:method-data}, which contains samples where object counts are spelled out in the captions. While $L_{CLIP}$ is calculated on all samples, the counting loss $L_{count}$ is calculated only on samples from $C$.
For each image-text pair ($i_{k}$, $t_{k}$) $\in C$, a counterfactual caption $t^{CF}_k$ is automatically created by swapping the number in the caption $t_{k}$ with a different random number (e.g., the caption ``five dogs'' can be counterfactualized with ``eight dogs''). 
At each training step, the triplets $(i_{k}$, $t_{k}$, $t^{CF}_{k})_{k=1}^N$ are then fed to CLIP's text and image encoders to obtain their embeddings $(ei_{k}$, $et_{k}$, $et^{CF}_{k})_{k=1}^N$.

Then, a contrastive loss $L_{count}$ is computed 
to enforce that the similarity score of the image is high with the original caption and low with the counterfactual caption:

\begin{equation}
    L_{count} = -\frac{1}{N}\sum_{k=1}^{N}\text{log}\frac{\text{exp}(ei_{k} \cdot et_{k})}{\text{exp}(ei_{k} \cdot et_{k}) + \text{exp}(ei_{k} \cdot et^{CF}_{k})}
\end{equation}

Since the original ground truth caption and counter-factual caption differ only by the number of objects specified in them, this loss encourages the model to learn the relationship between the specified spelled number and the number of the objects it refers to.

We use the negative samples only in the counting objective $L_{count}$, instead of adding them to the batch for the existing contrastive loss $L_{CLIP}$, in order to better weight their effect. 

\subsection{Implementation details}
\paragraph{Models.} We test our method with two classes of SOTA VLMs, BASIC~\cite{BASIC} and CLIP~\cite{clip}, in order to verify its robustness to different architectures. For CLIP, we experiment with both CLIP-B/32 and CLIP-L/14 configurations, as they are both widely used in recent work. For BASIC, we experiment with BASIC-S.

\paragraph{Training.} We finetune all models for $20K$ steps using a cosine schedule with an initial learning rate of $5e^{-6}$. We use a batch size of 32,768, where a fraction $p = \frac{1}{32}$ of each batch is dedicated to samples from the counting training set, and the rest are from large image-text datasets collected from the Web.  We use $\lambda = 1$ to weight the auxiliary loss, with a linear warm-up in the first 10,000 steps.

\begin{figure*}
    \centering
    \includegraphics[width=0.98\textwidth]{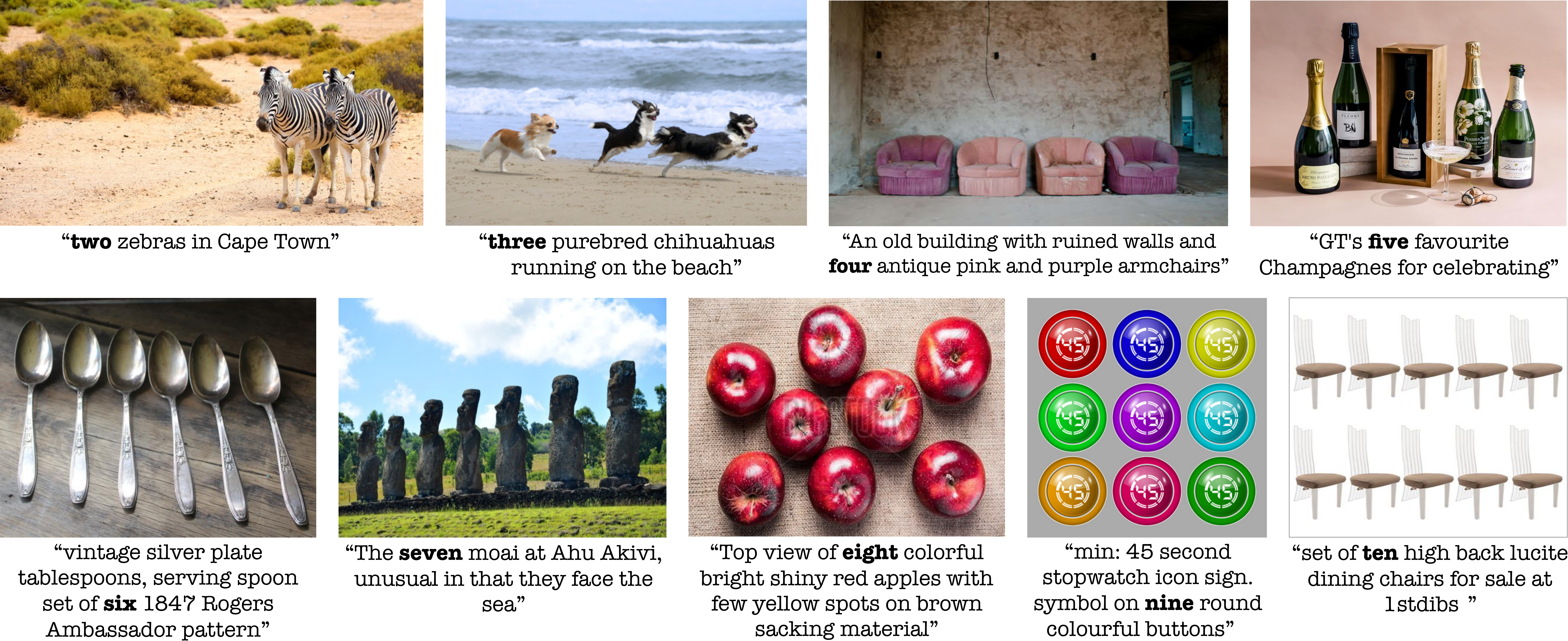}
  \vspace*{-0.1cm}
    \caption{\textbf{CountBench benchmark.} \it Sample images and their corresponding captions from our new {\it CountBench} object counting benchmark. This benchmark was automatically curated (and manually verified) from the publicly-available LAION-400M dataset. }
    \label{fig:countbench}
  \vspace*{-0.2cm}
\end{figure*}
\section{CountBench}
\label{sec:counbench}
We introduce a new object counting benchmark called {\it CountBench}, automatically curated (and manually verified) from the publicly available LAION-400M image-text dataset~\cite{schuhmann2021laion}. CountBench contains a total of 540 images containing between two and ten instances of a particular object, where their corresponding captions reflect this number.
This benchmark is used only for testing and is filtered from datasets which have no overlap with our training set $C$.

The images in {\it CountBench} were obtained by first running our automatic filtering method described in \cref{sec:method-data} on the entire LAION-400M dataset. This filtering produced over 158K images for the number $``two"$, but only around 100 for $``ten"$, demonstrating again the severe number imbalance we encountered with our training sets. After automatically balancing each number to 100-200 samples each, the entire dataset was manually verified to contain only pairs in which the spelled number in the caption matches the number of clearly visible objects in the image. The dataset was rebalanced after this stage, ending up with 60 image-text pairs per number $\in\set{``two",..,``ten"}$, 540 in total. Samples from the dataset can be seen in \cref{fig:countbench}.

It is worth noting that the higher the count is, the higher the proportion of CountBench images which contain relatively simplistic 2D collections of objects, as opposed to objects in a real-world scene. This bias exists in the training set as well, and seems to be a characteristic of web-scraped counting data in general.

We use the {\it CountBench} benchmark to evaluate the counting abilities of the models trained with our method in \cref{sec:exps}. These images are not used for training.

\begin{table*}
\begin{center}
\resizebox{0.98\linewidth}{!}{
\begin{tabular}{p{0.4\columnwidth}|ccccc|ccccc}
\toprule[1.5pt]
Dataset & \multicolumn{5}{c}{CLIP-B/32} & \multicolumn{5}{c}{BASIC-S} \\
& \underline A & \underline B & \underline C & \underline D & \underline E & \underline A & \underline B & \underline C & \underline D & \underline E \\
 & \bf Official & \bf Internal & \bf Ours & \bf Ours &\bf Ours & \bf Public & \bf Internal & \bf Ours & \bf Ours &\bf Ours \\ 
 & \bf Baseline & \bf Baseline & \bf \begin{small}(w/o $\boldsymbol{L_{count}}$)\end{small} & \bf \begin{small}(Naive Filtering)\end{small}  & & \bf Baseline & \bf Baseline & \bf \bf (w/o $\boldsymbol{L_{count}}$) & \bf \begin{small}(Naive Filtering)\end{small} & \\
\toprule
\bf{Accuracy $\uparrow$} & 31.67 & 32.94 & 44.26 & 49.81 & \bf{75.93} & 17.97 & 22.75 & 30.59 & 28.68 & \bf{69.02} \\
\bf{Mean deviation from the correct number $\downarrow$} & 1.53 & 1.44 & 0.97 & 1.28 & \bf{0.49} & 2.13 & 2.02 & 1.29 & 1.87 & \bf{0.64} \\
\bottomrule[1.5pt]
\end{tabular}
}
\end{center}
\vspace{-0.4cm}
\caption{{\bf Quantitative counting results.}
{ \it Top-1 zero-shot accuracy and the mean absolute distance between the predicted numbers and the true numbers on CountBench. We compare several configurations: (A) The official CLIP~\cite{clip} and BASIC~\cite{BASIC} models. (B) The official baselines finetuned on our internal curated data. (C) Models trained with our filtered counting set, without $L_{count}$ (D) Models finetuned with $L_{count}$ on a naively filtered counting set (E) Our method, which is significantly superior to all other configurations, both in accuracy and deviation from correct number.
}}
\label{tab:zeroshot} 
\end{table*}

\begin{table}
\begin{center}
\resizebox{0.98\linewidth}{!}{
\begin{tabular}{p{0.4\columnwidth}|ccc|ccc}
\toprule[1.5pt]
Dataset & \multicolumn{3}{c}{CLIP-B/32} & \multicolumn{3}{c}{BASIC-S} \\
 & \bf Official & \bf Internal & \bf Ours & \bf Public & \bf Internal &\bf Ours \\ 
 & \bf Baseline & \bf Baseline & & \bf Baseline & \bf Baseline & \\
\toprule
ImageNet & 62.93 & 64.97 & 64.06 & 59.70 & 61.96 & 61.18 \\
CIFAR10 & 63.91 & 61.00 & 60.65 & 76.22 & 84.69 & 84.05 \\
CIFAR100 & 33.10 & 32.49 & 33.56 & 45.35 & 56.80 & 55.89 \\
Caltech101 & 75.99 & 82.50 & 82.36 & 78.16 & 81.03 & 81.05 \\
EuroSAT & 45.23 & 41.66 & 37.69 & 28.39 & 45.82 & 45.97 \\
Food101 & 83.08 & 80.72 & 80.53 & 77.08 & 77.80 & 77.06 \\
ImageNetA & 31.85 & 30.85 & 29.81 & 17.65 & 22.55 & 21.68 \\
ImageNetR & 69.38 & 70.17 & 70.30 & 67.11 & 67.68 & 66.95 \\
ImageNetV2 & 55.65 &56.56 & 56.62 & 52.22 & 54.35 & 53.60 \\
Oxford Pets & 87.35 & 87.74 & 87.41 & 80.62 & 85.15 & 84.87 \\
Oxford Flowers & 66.14 & 65.73 & 67.39 & 64.74 & 66.40 & 65.90 \\
\bottomrule[1.5pt]
\end{tabular}
}
\end{center}
\vspace{-0.4cm}
\caption{{\bf Zero-shot accuracy on common benchmarks.} \it We compare the zero-shot accuracy of our method and baselines on a variety of popular benchmarks. As can be seen, our method preserves the performance of the original model.}
\label{tab:zeroshot_common_benchmarks} 
\end{table}

\begin{figure}
    \centering
    \includegraphics[width=\linewidth]{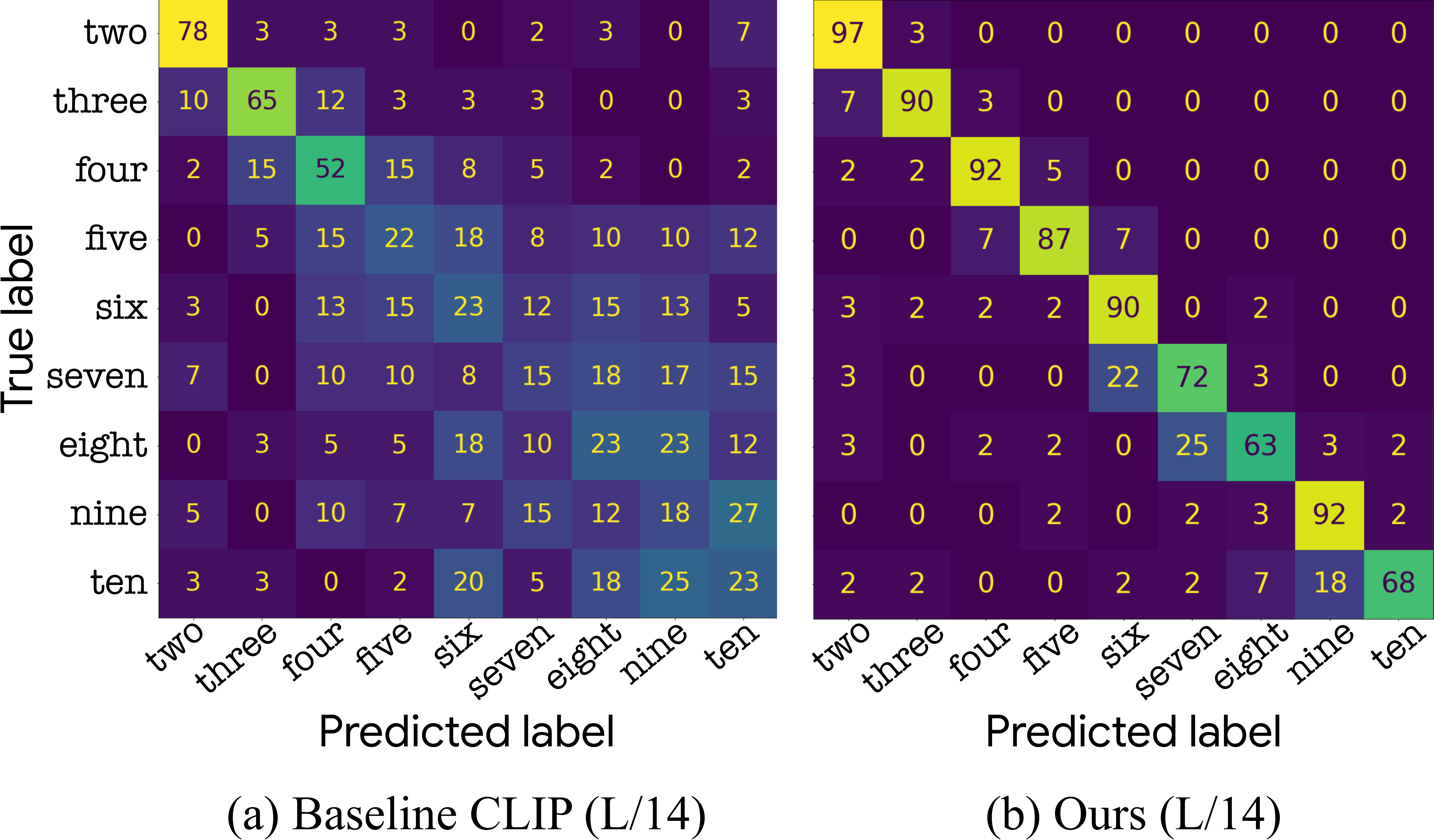}
  \vspace*{-0.3cm}
    \caption{{\bf Confusion matrices on CountBench.} \it Classification accuracy on our new counting benchmark, {\it CountBench}, broken down into confusion matrices for the public CLIP ViT-L/14 (a), and our improved CLIP ViT-L/14 model (b), demonstrating clear quantitative superiority of our model. 
    }
    \label{fig:confmat-counting}
  \vspace*{-0.3cm}
\end{figure}
\section{Experiments}
\label{sec:exps}

We thoroughly evaluate our method, both quantitatively and qualitatively,  on object counting-related tasks using our {\it CountBench} benchmark. We further validate that the performance of our finetuned counting-aware models on a variety of \emph{general} zero-shot classification benchmarks is retained~\cite{imagenet, imagenetv2, imagenet-r, Cifar10, cifar100, caltech, EuroSAT, Food101, oxforspets, oxfordflowers, imagenetA}. To gain a better understanding of our models, we show visualizations of text-image relevancy maps, along with per-word relevancy scores, demonstrating that our model indeed attends to the number of objects in the image and text. Finally, we apply our model to text-to-image retrieval and generation, producing specific numbers of objects more reliably than baseline models.

\subsection{Zero-shot counting accuracy} %
\label{sec:zeroshotacc}

We evaluate our models and baselines on {\it CountBench} on the task of classifying the number of objects in an image in a zero-shot manner. For each image in {\it CountBench} we augment the existing caption with eight other possible captions by replacing the number in its caption with all the numbers $\in \{``two", \dots,``ten"\}$, and calculate the similarity score between the image and each of the nine captions. The number in the caption that obtains highest similarity score with the image is considered the predicted number.

\Cref{tab:zeroshot} reports the results of this evaluation on two prominent contrastive VLMs: CLIP-B/32 and BASIC-S.
We report both the counting accuracy (selection of the correct number) and the mean deviation of the models' predictions from the correct numbers.
For each of the architectures, we compare our model (configuration {\it E}) with two baseline configurations: ({\it A}) the official baseline model, and ({\it B}) the baseline model finetuned on our general text-image dataset used in our implementation, with the standard contrastive loss. Comparing the performance of these configurations allows us to quantify the effect of using our own large-scale text-image dataset, which differs from the original unpublished data the models were trained on.

As can be seen, our method ({\it E}) achieves significantly superior counting accuracy compared to the baselines ({\it A, B}). Our counting-aware CLIP and BASIC models achieve $2\textrm{--}3\times$ higher counting accuracy than their corresponding baselines and more than $3\times$ lower mean deviation from the correct number. 

\cref{tab:zeroshot} also contains an ablation of the two components of our method: filtering a counting training set and finetuning with an additional loss $L_{count}$.
Models with configuration {\it C} are finetuned on the filtered subset with no counting loss. The large gap in accuracy on {\it CountBench} between configurations {\it C} and {\it E} shows the importance of our loss for the improvement in counting abilities.
Models with configuration {\it D} are finetuned with the counting loss $L_{count}$ on an alternative counting subset, which consists of all the samples that contain spelled numbers  $\in\set{``two",..,``ten"}$ without additional filtering.
The significant difference in counting accuracy between configurations {\it D} and {\it E} demonstrates the importance of our restrictive filtering pipeline, as both configurations are finetuned with $L_{count}$ over the samples from a dedicated counting training set. As can be seen in ~\cref{tab:zeroshot}, while the naively filtered data does improve performance over a baseline trained without a dedicated counting subset, the obtained results are still significantly lower than those produced by our model. We attribute this gap in performance to mislabeled training samples in the naively filtered data, which are absent from our counting training set $C$ due to our filtering pipleine.

Confusion matrices for the counting evaluation described above are shown in \cref{fig:confmat-counting}. For this experiment, we compare a CLIP-L/14 model finetuned with our method against the public CLIP-L/14 model checkpoint. As can be seen, our improved CLIP model is significantly superior to the baseline across all numbers. Also evident is a dropoff in accuracy for some higher numbers, as a result of their significantly lower representation in the training data (detailed above in Sec.~\ref{sec:method-data}).

\paragraph{Performance on common non-counting classification tasks}
To verify that our counting-aware models preserve the powerful image-text representation capabilities of the original models, we evaluate the zero-shot performance of our models on a variety of common classification benchmarks. \Cref{tab:zeroshot_common_benchmarks} reports the
zero-shot accuracy of our counting aware models against the baselines (corresponding to configurations {\it A, B} in \cref{tab:zeroshot}).
As can be seen, our models maintain similar overall accuracy. Also, comparing the official baseline and the internal baseline indicates that finetuning the models on our general text-image datasets leads to only a slight shift in the accuracy of the models on common benchmarks. 

\paragraph{Hyperparameters of our method}
Our method introduces two additional hyperparameters: the portion $p \in[0,1]$ of the batch size dedicated to the counting subset, and the weight $\lambda$ of our counting loss $L_{count}$. We empirically chose $p=\frac{1}{32}$ and $\lambda = 1$, since higher values tend to overfit to the counting subset. \cref{tab:p_ablation} contains an ablation our choice of $p$, and \cref{tab:aux_ablation} compares the results of models trained with different weightings $\lambda$ of $L_{count}$.

\subsection{Count-based image retrieval}
\label{retrieval}
We consider the task of text-to-image retrieval where the text explicitly describes the desired count of objects. 
To obtain a diverse dataset that consists of varied numbers of objects, yet facilitates retrieval in reasonable time, we split the public LAION-400M dataset~\cite{schuhmann2021laion} into coarse categorical subsets by filtering samples where the caption contains a certain word (e.g., ``dogs'', ``animals'', ``cars''), and perform retrieval on each of these subsets separately. 

\begin{figure*}
    \centering
    \includegraphics[width=0.9
\textwidth]{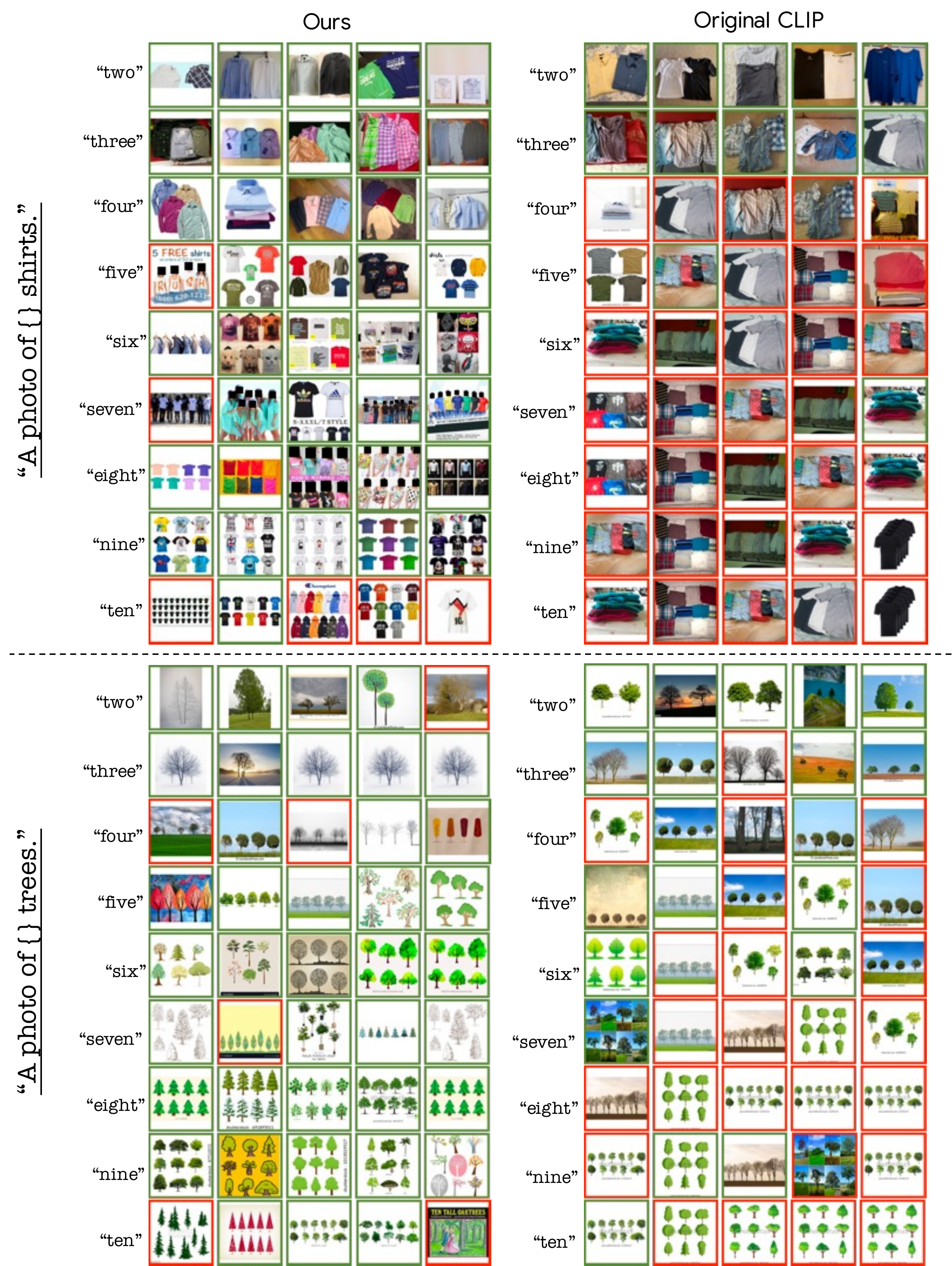}
  \vspace*{-0.2cm}
    \caption{\textbf{Top-5 count-based image retrieval} \it Text-to-image retrieval results for different counts of objects (images that match the caption are colored in green, and images that don't match it are colored in red). 
    The images are ordered according to their similarity scores, such that the images with the highest scores are in the left column and the images with the fifth-highest scores are in the right column. 
    As can be seen, the retrieval results of our model are significantly more accurate than the original CLIP model, which often fails when the requested number is higher than three.} 
    \label{fig:retrieval_fig}
  \vspace*{-0.3cm}
\end{figure*}

For each category, we use the caption ``a photo of $n$ \{objects\}" where $n \in \{``two", .., ``ten"\}$ (e.g. ``a photo of six dogs").
For each caption, we retrieve the five images in the dataset that are predicted by the model to be most similar to the caption.
Note that since there are no ground truth labels for the counts of objects, we present qualitative results.
\cref{fig:retrieval_fig} shows the retrieved images using the original CLIP model and our counting-aware CLIP model. As can be seen, when the requested number is larger than three, the images retrieved by the baseline model often depict arbitrary numbers of objects. Additionally, the baseline often retrieves the same images for several different requested numbers. This further implies that the baseline model mostly focuses on the existence of the described object in the image, and ignores the number in the caption. In contrast, our results depict accurate object counts in most cases.

\subsection{Relevancy map visualization}
\begin{figure}
    \centering
    \includegraphics[width=\linewidth]{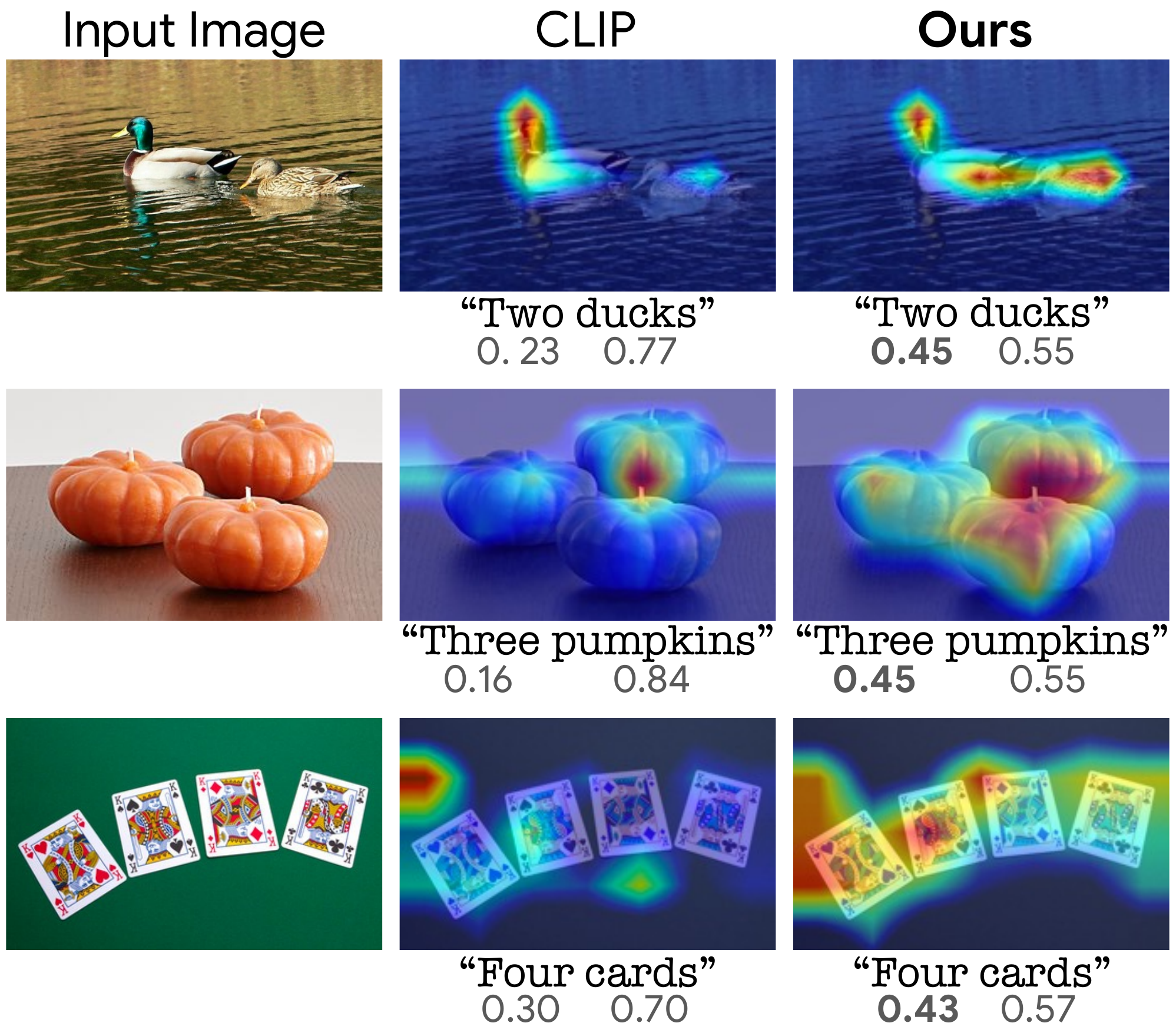}
    \vspace*{-0.5cm}
\caption{\textbf{Relevancy map of both image and text.}  \it {Visualization of the relevancy scores of both image and text, which represent, for each patch in the image and token in the text, how important it is to the prediction. Using our improved CLIP model, the relevancy of the number (e.g., ``four") in the text is increased. In addition, the model focuses on areas in the image that are relevant for counting.}}
    \label{fig:relevancy_map_count}
  \vspace*{-0.3cm}
\end{figure}
\begin{figure*}
    \centering
    \includegraphics[width=0.98\textwidth]{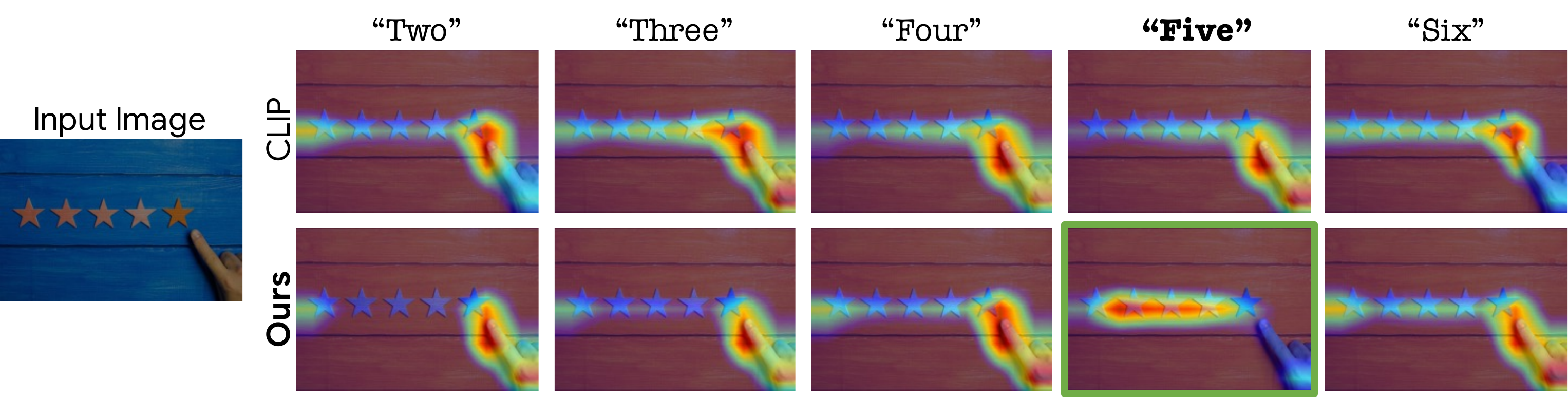}
  \vspace*{-0.2cm}
    \caption{\textbf{Relevancy maps for similarity between the image and different numbers.} {\it We compare the CLIP relevancy map of the input image with text prompts of several numbers (i.e. two to six) for both the baseline CLIP model and for our model. Our CLIP model focuses on the five stars when calculating the similarity with the prompt ``five''. }}
    \label{fig:relevancy_map_stars}
  \vspace*{-0.3cm}
\end{figure*}
To gain a better understanding of what our model learns, we use an explainability method to visualize the reasoning of the model.
For each image-caption pair, we refer to the cosine similarity of their CLIP embeddings as their similarity score. This score should be high for a pair that CLIP considers matching and low for non-matching images and texts. We use the method of Chefer et al.~\cite{Chefer2021GenericAE} to obtain relevancy maps, which consist of relevancy scores for every patch in the image and every token in the text.

The relevancy scores indicate the importance of different parts of the text and image in predicting the similarity score of the model. ~\cref{fig:relevancy_map_count} displays the relevancy maps of several image-text pairs. Note that the relevancy scores of the text are normalized to sum to 1. Examining the relevancy maps of the text, it is apparent that the relevancy score of the spelled number in the caption is significantly higher than the baseline model, which suggests that our model concentrates more on the mentioned number than the original one. Additionally, examining the relevancy maps of the images, it is evident that our model focuses on all pertinent objects in the image, whereas the original model primarily identifies a single instance of the described object.

To verify that our model does not simply attend to {\it all} objects that appear in the image, we examined the relevancy maps in ~\cref{fig:relevancy_map_stars} using negative text prompts (\ie the text ``three'' when there are five elements in the image). Our model focuses only on relevant objects when the correct number is used, unlike the baseline CLIP model that highlights all object types in the image. This demonstrates that our model learns to associate the spelled number in the caption with the suitable number of objects, and does not exploit shortcuts or undesired content.

\subsection{Text-to-image generation}
In order to demonstrate the effectiveness of our fine-tuned model on downstream image generation tasks, we train an Imagen~\cite{imagen} model, conditioned on the textual embeddings of a pretrained CLIP-L/14, and another Imagen model conditioned on our counting-aware CLIP-L/14 model.
To compare our model and the baseline, we synthesize $12$ samples for each textual prompt in the counting category of the DrawBench benchmark~\cite{imagen}. For each sample, we check whether or not it contains the requested number of objects, as stated in its prompt.
We report the total binary accuracy in \cref{tab:draw_bench}. 

Since the highest number specified in DrawBench for a given object is five, we obtain an additional set of prompts by generating all possible combinations of the form ``\{$number$\} \{$class$\}.", where $number \in \set{``two",...,``ten"}$ and $class$ is one of the classes in CIFAR10 (e.g., ``dog" and ``car"). 
Since the amount of training samples that contain the numbers $2-6$ greatly exceeds those of higher values, we additionally report the accuracy for the textual prompts containing numbers within the range of $2-6$. As shown in \cref{tab:draw_bench}, our finetuning approach leads to a $1.5-2\times$ improvement in the ability to reliably generate specific counts of objects.

\subsection{Limitations}
\begin{figure}
    \centering
    \includegraphics[width=\linewidth]{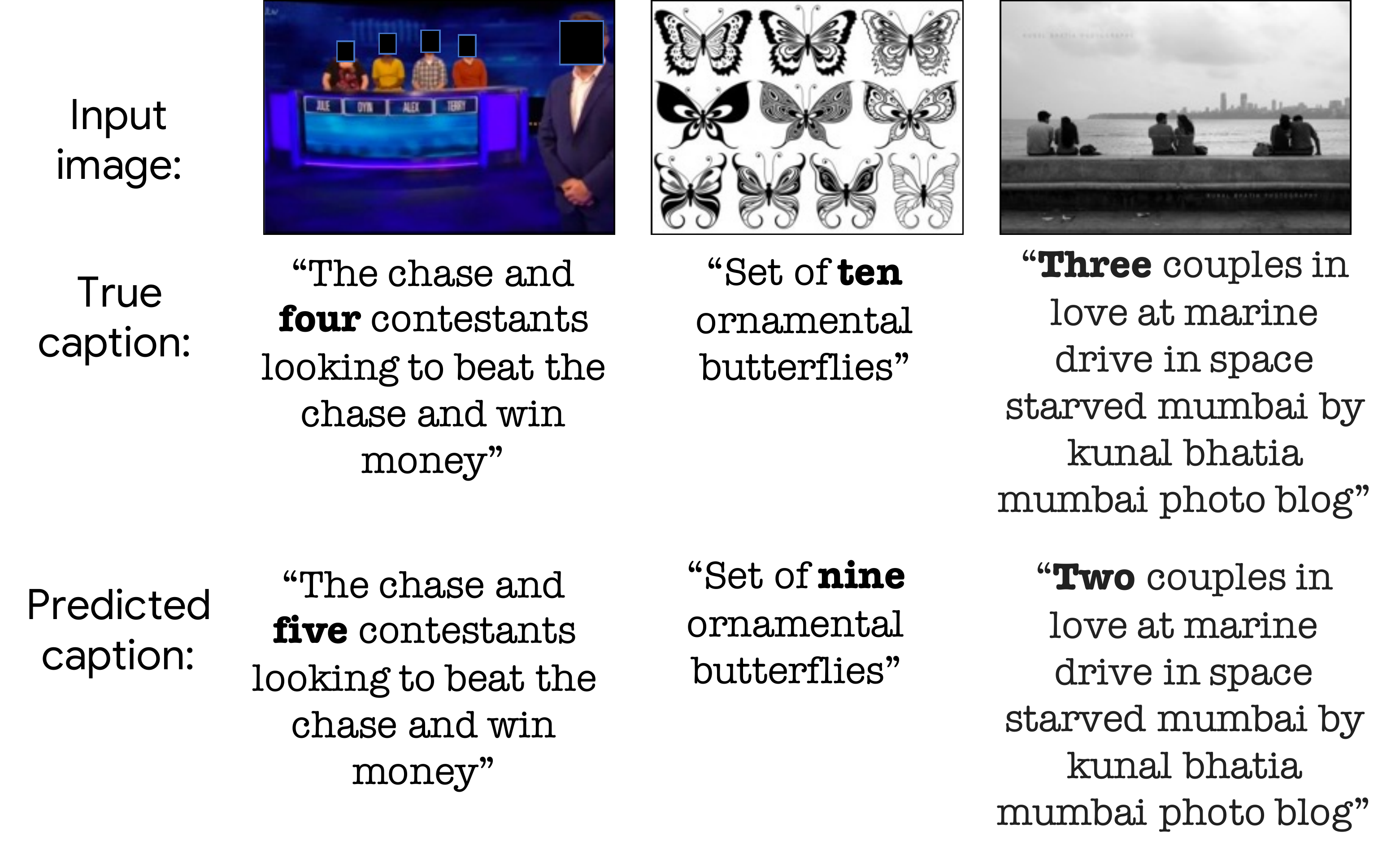}
    \vspace*{-0.5cm}
\caption{\textbf{Exemplary failure cases of our method.} \emph{The model struggles with captions that require prior knowledge, images oriented in grids typical to other counts, or numbers paired with another word that specifies amounts such as ``couple".}}
    \label{fig:limitations}
  \vspace*{-0.3cm}
\end{figure}

First and foremost, our method is limited by the insufficient existence of training data with images containing multiple instances of an object, along with a corresponding caption that correctly spells out this information. The effect of this data scarcity on our method increases with larger numbers (7, 8, etc.) as people tend to use ``a group of" or ``many" for large numbers of objects, instead of gruelingly counting them. Furthermore, many of the correct training pairs with higher numbers that do exist, contain relatively simplistic 2D collections of objects, as opposed to objects in a real-world scene (see \cref{fig:countbench}), and can explain weaker model performance on in-the-wild images containing a larger number of objects.
In addition, our method teaches CLIP to count only until ten, while generalization to numbers greater than ten is unclear. We did not evaluate on these numbers due to lack of data. Examplary failure cases can be found in \cref{fig:limitations}.

\begin{table}
\begin{center}
\resizebox{0.95\linewidth}{!}{
\begin{tabular}{p{0.4\columnwidth}|ccc}
\toprule[1.5pt]
Dataset & \bf $p=\frac{1}{32}$ & \bf $p=\frac{1}{8}$ & \bf $p=\frac{1}{4}$\\[4pt]
\toprule
\begin{small}\bf{CountBench}\end{small} & \begin{small}\bf{75.93}\end{small} & \begin{small}70.19\end{small} & \begin{small}69.81\end{small} \\
\midrule
\begin{small}ImageNet\end{small} & \begin{small}64.06\end{small} & \begin{small}64.11\end{small} & \begin{small}63.96\end{small} \\
\begin{small}CIFAR10\end{small} & \begin{small}60.65\end{small} & \begin{small}61.69\end{small} & \begin{small}63.04\end{small}\\
\begin{small}CIFAR100\end{small} &\begin{small}33.56\end{small} & \begin{small}33.74\end{small} & \begin{small}34.01\end{small}\\
\begin{small}Caltech101\end{small} & \begin{small}82.36\end{small} & \begin{small}83.58\end{small} & \begin{small}83.51\end{small}\\
\begin{small}EuroSAT\end{small} & \begin{small}37.69\end{small} & \begin{small}39.07\end{small} & \begin{small}41.56\end{small}\\
\begin{small}Food101\end{small} & \begin{small}80.53\end{small} & \begin{small}80.59\end{small} & \begin{small}80.80\end{small}\\
\begin{small}ImageNetA\end{small} & \begin{small}29.81\end{small} & \begin{small}30.84\end{small} & \begin{small}30.60\end{small}\\
\begin{small}ImageNetR\end{small} & \begin{small}70.30\end{small} &\begin{small}70.15\end{small} & \begin{small}69.98\end{small}\\
\begin{small}ImageNetV2\end{small} & \begin{small}56.62\end{small} & \begin{small}56.54\end{small} & \begin{small}56.37\end{small}\\
\begin{small}Oxford Pets\end{small} & \begin{small}87.41\end{small} & \begin{small}87.14\end{small} & \begin{small}86.64\end{small}\\
\begin{small}Oxford Flowers\end{small} & \begin{small}67.39\end{small} & \begin{small}67.21\end{small} & \begin{small}67.91\end{small} \\
\bottomrule[1.5pt]
\end{tabular}
}
\end{center}
\vspace{-0.3cm}
\caption{\bf Ablation of hyperparameter $\boldsymbol{p}$. {\it $p$ denotes the fraction of the batch size dedicated to samples from the counting subset. As the subset is significantly smaller than the entire curated dataset we found that large values for $p$ lead to overfitting.}}
\vspace*{-0.3cm}
\label{tab:p_ablation} 
\end{table}

\begin{table}
\begin{center}
\resizebox{0.95\linewidth}{!}{
\begin{tabular}{p{0.4\columnwidth}|cccc}
\toprule[1.5pt]
Dataset & \bf $\lambda = 0.1$ & \bf $\lambda = 1$ & \bf $\lambda = 5$ & \bf $\lambda = 10$\\
\toprule
\bf{CountBench} & 69.44 & \bf{75.93} & 73.15 & 72.59 \\
\midrule
ImageNet & 64.50 & 64.06 & 63.84 & 63.53 \\
CIFAR10 & 63.20 & 60.65 & 63.79 & 63.82 \\
CIFAR100 & 34.51 & 33.56 & 35.35 & 34.15 \\
Caltech101 & 84.39 & 82.36 & 81.82 & 81.76 \\
EuroSAT & 39.48 & 37.69 & 39.93 & 42.20\\
Food101 & 80.73 & 80.53 & 80.33 & 79.98\\
ImageNetA & 31.67 & 29.81 & 29.55 & 29.45\\
ImageNetR & 70.92 & 70.30 & 69.87 & 69.77 \\
ImageNetV2 & 56.70 & 56.62 & 56.30 & 56.09\\
Oxford Pets & 87.65 & 87.41 & 87.79 & 86.97 \\
Oxford Flowers & 67.00 & 67.39 & 65.33 & 65.90\\
\bottomrule[1.5pt]
\end{tabular}
}
\end{center}
\vspace*{-0.4cm}
\caption{{ \bf Ablation of the auxilary loss weight $\lambda$} \it We ablate different weights for the auxilary loss. We found $\lambda = 1$ to work best, as lower values lead to suboptimal results and higher values cause overfitting.}
\vspace*{-0.3cm}
\label{tab:aux_ablation} 
\end{table}

\begin{table}
\begin{center}
\resizebox{0.95\linewidth}{!}{
\begin{tabular}{p{0.25\columnwidth}ccc}
\toprule[1.5pt]
 & \bf DrawBench & \bf CIFAR10 (2-6) & \bf CIFAR10 (2-10) \\
\toprule
Baseline CLIP & 24.12 & 34.33 & 20.00  \\
\midrule
Ours & \bf{40.35}  & \bf{68.83} & \bf{50.18} \\
\bottomrule[1.5pt]
\end{tabular}
}
\end{center}
\vspace{-0.3cm}
\caption{{\bf Text-conditioned image generation evaluation.}
{ \it We compare an Imagen model trained with the official CLIP against a model trained with our counting-aware CLIP model on the task of generating images with a specific number of objects. For each textual prompt we generate $12$ images and tag each result as correct or incorrect based on whether it matches the specified number of objects. The table reports the binary accuracy of this evaluation. Our counting-aware CLIP leads to a $1.5-2\times$ improvement in the ability of Imagen to reliably generate specific counts of objects.}
}
\label{tab:draw_bench} 
\end{table}

\section{Conclusions and future work}
This work presents the first method to enhance CLIP with counting abilities, which is an essential step towards enabling more accurate retrieval and generation of detailed texts. Using a carefully designed filtering pipeline, we are able to obtain a clean counting subset from datasets collected from the internet, on which we perform counting-focused hard-negative augmentation. An additional loss is applied that encourages CLIP to understand object counting, in order to successfully separate false captions from images. In addition, we introduce a new counting benchmark, {\it CountBench}, which we plan to release publicly, that contains in-the-wild images and captions where the number of specific objects in the image is detailed in the caption. We hope this benchmark will encourage more research in this direction in the future. Applying our improved CLIP to the task of image generation is shown to improve reliability of producing specific counts of objects.

 While the method is not specific to counting, and can also be applied on other compositional concepts that VLMs fail to learn, we focus on counting, as it is the most unambiguous to define and evaluate, and allows us to disentangle the model's understanding of the concept from textual or visual ambiguities. The extension of this method to other compositional concepts such as spatial positioning of objects, active vs. passive verbs, etc, remains for future work. 

\paragraph{Societal impact}
Our work aims to improve the discriminative representation of numbers within vision-language models. Those capabilities can be used to improve downstream applications such as text-to-image synthesis and text-based image editing. These could be used by malicious parties for synthesizing fake imagery to mislead viewers. It should be noted, however, that our contribution to the improvement of these models is for the specific application of generating a specific number of objects in an image, and should not be considered a novel image generation method in itself. As with other image generation work, mitigation of malicious use depends on further research on identification of synthetically edited or generated content.

\paragraph{Acknowledgements}
We thank Hieu Pham for his technical guidance and insightful feedback.

{\small
\bibliographystyle{ieee_fullname}
\bibliography{egbib}
}

\clearpage

\appendix

\begin{table*}
\begin{center}
\resizebox{0.95\linewidth}{!}{
\begin{tabular}{p{0.25\columnwidth}cccccc}
\toprule[1.5pt]
 & \multicolumn{2}{c}{\bf \underline{prompts from DrawBench}} & \multicolumn{2}{c}{\bf \underline{CIFAR-10 class labels (numbers 2-6)}} & \multicolumn{2}{c}{\bf \underline{CIFAR-10 class labels (numbers 2-10)}} \\[4pt]
  & \bf Accuracy $\uparrow$ & \bf MAE $\downarrow$ & \bf Accuracy $\uparrow$ & \bf MAE $\downarrow$ & \bf Accuracy $\uparrow$ & \bf MAE $\downarrow$ \\
\toprule
Baseline CLIP & 24.12 & 0.94 & 34.33 & 0.78 & 20.00 & 3.32 \\
\midrule
Ours& \bf{40.35}  & \bf{0.81} & \bf{68.83} & \bf{0.38} &\bf{50.18} & \bf{1.09} \\
\bottomrule[1.5pt]
\end{tabular}
}
\end{center}
\vspace{-0.3cm}
\caption{{\bf Text-conditioned image generation evaluation.}
{ \it We compare an Imagen model conditioned on the baseline CLIP against a model trained with our counting-aware CLIP model. For each textual prompt within the DrawBench counting category, we generate $12$ images and tag whether or not they match the textual prompt w.r.t. the number of the requested objects. 
}}
\label{tab:draw_bench_ext} 
\end{table*}

\section{Image generation experiments}
\label{app:image_generation}
In this section, we provide further details of the text-conditioned image generation experiments mentioned in Sec. 5.3 of the main text (``Text-to-image generation"). 

\subsection{Experiment settings}
We train Imagen~\cite{imagen} models for $500K$ steps with a batch size of $512$ on $64$ TPUv4 chips. 
We employ the Adam~\cite{Kingma2015AdamAM} optimizer with a cosine learning rate schedule where the peak learning rate is $1e-4$, as done in the Imagen paper.
We remove the central-cropping augmentation used in the original model, as it can mislead the model when objects are cropped out of the image while still described in the caption. Instead, we pad the input images before resizing them to the $64\times64$ resolution. We additionally set a small portion ($3\%$) of the training batch to contain samples from our counting set while training our Imagen model.  As the number of objects in the image determined by the $64\times64$ model, we do not train the $256\times 256$ and $1024\times1024$ super-resolution models. Instead, we use the existing super-resolution models to generate $1024\times1024$ resolution images.

\subsection{Text prompts used for evaluation}
\begin{figure*}[!b]
    \centering
    \includegraphics[width=0.90\linewidth]{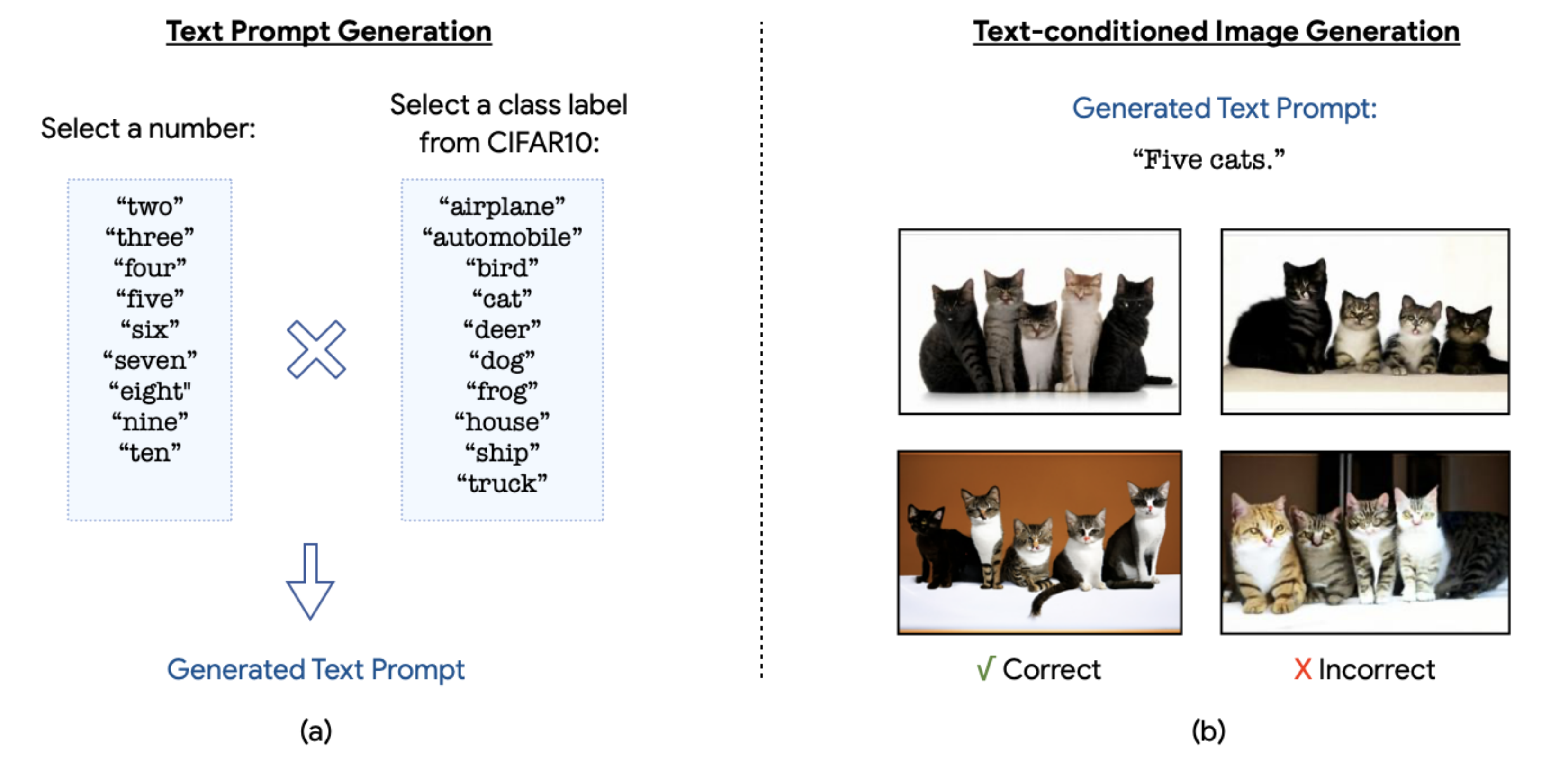}
  \vspace*{-0.2cm}
    \caption{{\bf An overview of the prompt generation pipeline.} \it As detailed in \cref{app:image_generation}, we create a set of captions containing the numbers $``two",..,``ten"$ and the class labels from CIFAR-10. (a) Each combination of number and class label is used to create a text prompt (b) We use the Imagen models
    to generate images based on the text prompt and measure accuracy and MAE.
    }
    \label{fig:geneval}
  \vspace*{-0.2cm}
\end{figure*}

We evaluate two Imagen models: one trained with the baseline CLIP and another conditioned on our counting-aware CLIP as a text backbone, on two predefined sets of textual prompts: 
\begin{enumerate}
  \item The prompts in the ``Counting" category of DrawBench~\cite{imagen}. Drawbench is designed to test text-to-image generative models on challenging prompts, and contains different categories of challenging scenarios. One of these categories is counting, which contains 19 prompts that describe numbers of objects, for example: {\it ``Two dogs on the street"}. The specified numbers range between one to five.
  \vspace{-0.18cm}
  \item To evaluate the model on captions with larger numbers of objects, we construct an additional set of prompts by creating all possible combinations of $``\{number\}~\{label\}"$ where $number \in \set{``two", .., ``ten"}$ and $label$ is one of the class labels of CIFAR-10 dataset~\cite{Cifar10}. This process, which is illustrated in \cref{fig:geneval}, results in $90$ distinct text prompts.
\end{enumerate}

\vspace{0.2cm}
\subsection{Evaluation protocol}
For each text prompt, we generate $12$ images using a DDPM sampler~\cite{ho2020denoising} with different random seeds, resulting in a total of $1296$ images. We manually count the number of instances of the requested object contained in each generated image, and compare it to the number specified in the prompt. 
For prompts that contain two specified numbers, such as {\it ``Three cats and one dog sitting on the grass"},  we follow the standard DrawBench procedure and consider successful generation as images containing the correct amount of both object categories.

\subsection{Results}
The results of our evaluation are reported in Tab.~4 in the main text, and again in \cref{tab:draw_bench_ext}, with the additional metric of mean absolute error (MAE) of the number of objects in the generated image, as compared to the number specified in the prompt. As can be seen, the results of the Imagen model trained with our counting-aware CLIP are around $2\times$ better than the results of the Imagen model trained with the baseline CLIP. Additional analysis is presented in \cref{tab:cifar_mae}, where we report MAE for each requested number separately. Evidently, as the numbers increase, so do the errors, for both our model and the baseline. However, even when it is wrong, our model clearly comes much closer to the desired number than the baseline. Our model has a drop in MAE for the label $``nine"$, which we attribute to the fact that many of the images with this label in the data are spatially organized in a grid-like structure, which makes them easier to learn.

\begin{figure}[]
    \centering
    \includegraphics[width=1\linewidth]{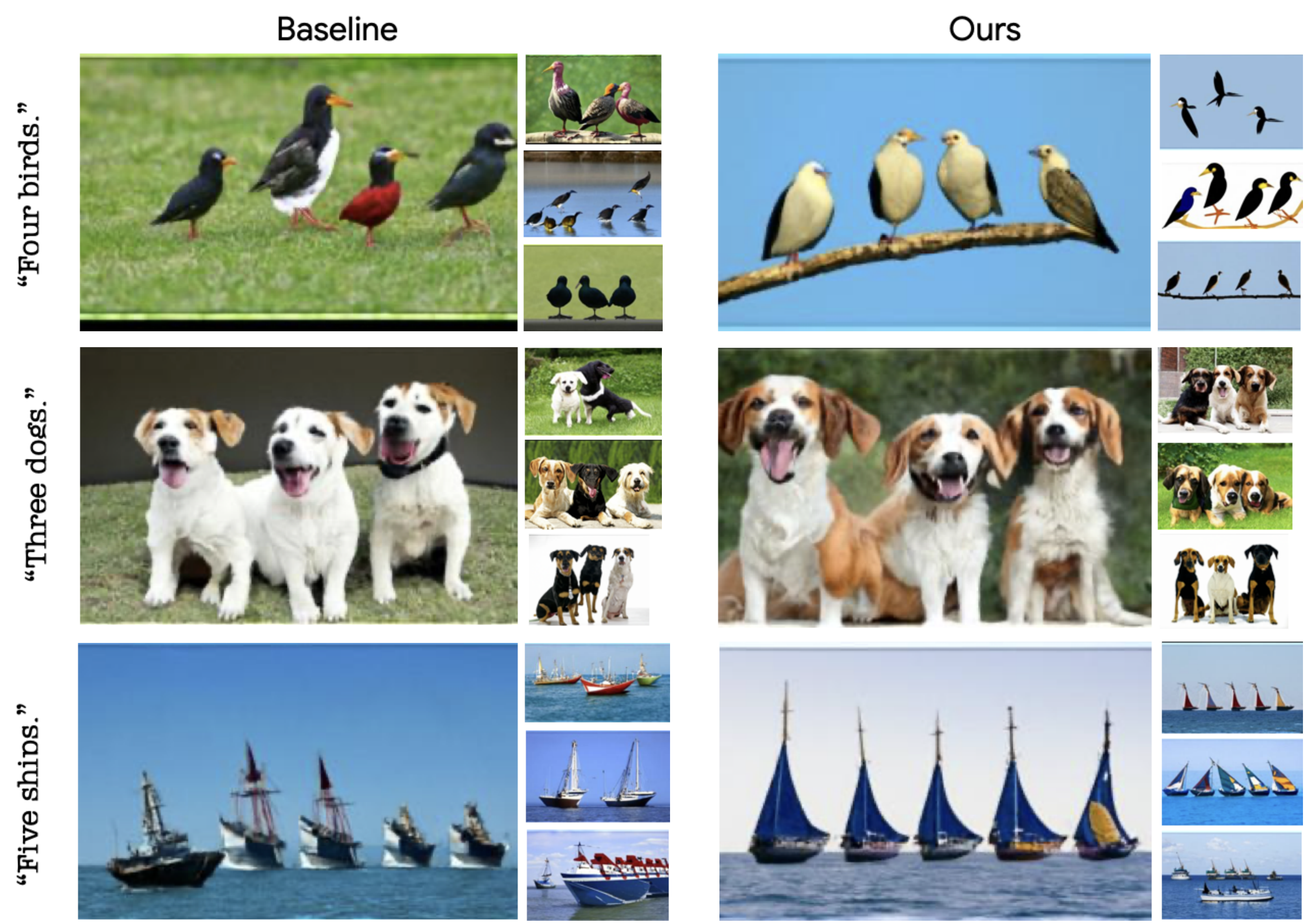}
  \vspace*{-0.2cm}
    \caption{\bf{Qualitative comparison of generated images.} \it{We show random images generated using textual prompts from the CIFAR-10 generated captions (see \cref{fig:geneval}). }}
    \label{fig:comparison}
  \vspace*{-0.2cm}
\end{figure}

\begin{table}
\begin{center}
\resizebox{0.95\linewidth}{!}{
\begin{tabular}{p{0.35\columnwidth}cc}
\toprule[1.5pt]
& \bf Baseline CLIP& \bf Ours \\
\toprule
$``two"$ & 0.23  & \bf 0.12 \\
$``three"$ &0.61 & \bf 0.25 \\
$``four"$ & 1.4 & \bf 0.52\\
$``five"$ & 1.96 & \bf 0.39\\
$``six"$ & 2.79 & \bf 0.44 \\
$``seven"$ & 4.33 & \bf 1.78\\
$``eight"$ & 4.58 & \bf 1.83  \\
$``nine"$ & 6.68 & \bf 1.06 \\
$``ten"$ & 7.31 & \bf 3.47\\

\bottomrule[1.5pt]
\end{tabular}
}
\end{center}
\vspace{-0.3cm}
\caption{\bf{MAE of the number of generated objects.} \it{For each generated image, we measure the mean absolute error (MAE) between the generated number of objects in the image to the number requested by the textual prompt. This table corresponds to the rightmost column in Tab.~\ref{tab:draw_bench_ext}.}}
\label{tab:cifar_mae} 
\end{table}

\section{Additional text-conditioned image generation examples}
\label{app:generation_results}
\begin{figure*}[b!]
    \centering
    \includegraphics[width=0.95\linewidth]{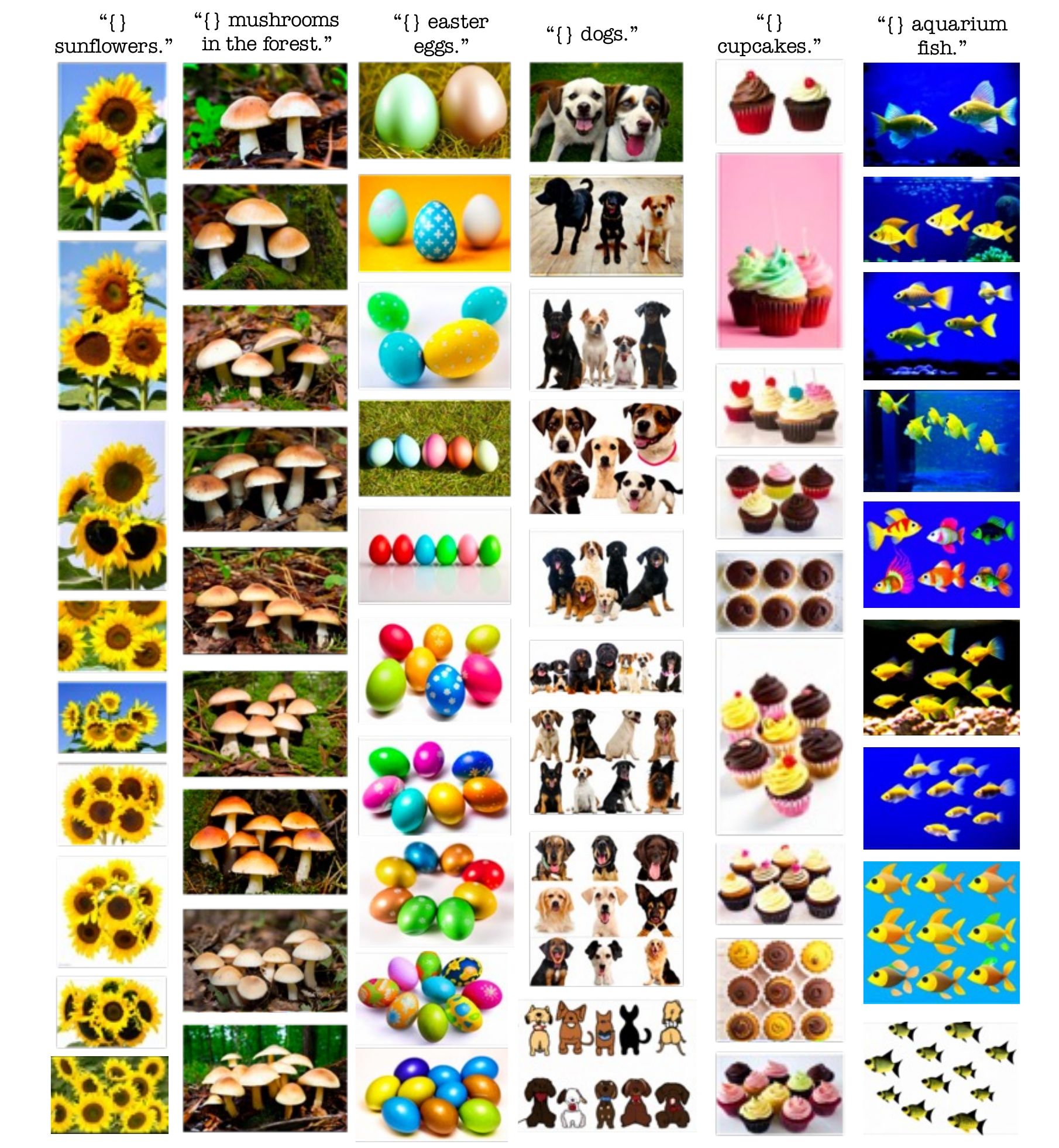}
  \vspace*{-0.2cm}
    \caption{{\bf Images generated with the Imagen model trained with our counting-aware CLIP}. \it For each of the caption templates at the top we inject numbers between ``two" and ``ten". The images generated conditioned on these prompts are ordered according to the injected number, such that the top-most images contain two objects and the bottom images contain ten objects.}
    \label{fig:generation_samples}
  \vspace*{-0.2cm}
\end{figure*}

Fig.~\ref{fig:comparison} shows a qualitative comparison between images generated with our method and the baseline. As can be observed, while the baseline model occasionally generates the correct number of objects, our method produces specific counts of objects more reliably.
Fig.~\ref{fig:generation_samples} presents additional images generated with the Imagen model trained with our counting-aware CLIP for prompts that specify the number of objects.

\section{Visualization of \textbf{\textit{CountBench}} benchmark}
\begin{figure*}[t!]
    \centering
    \includegraphics[width=0.90\linewidth]{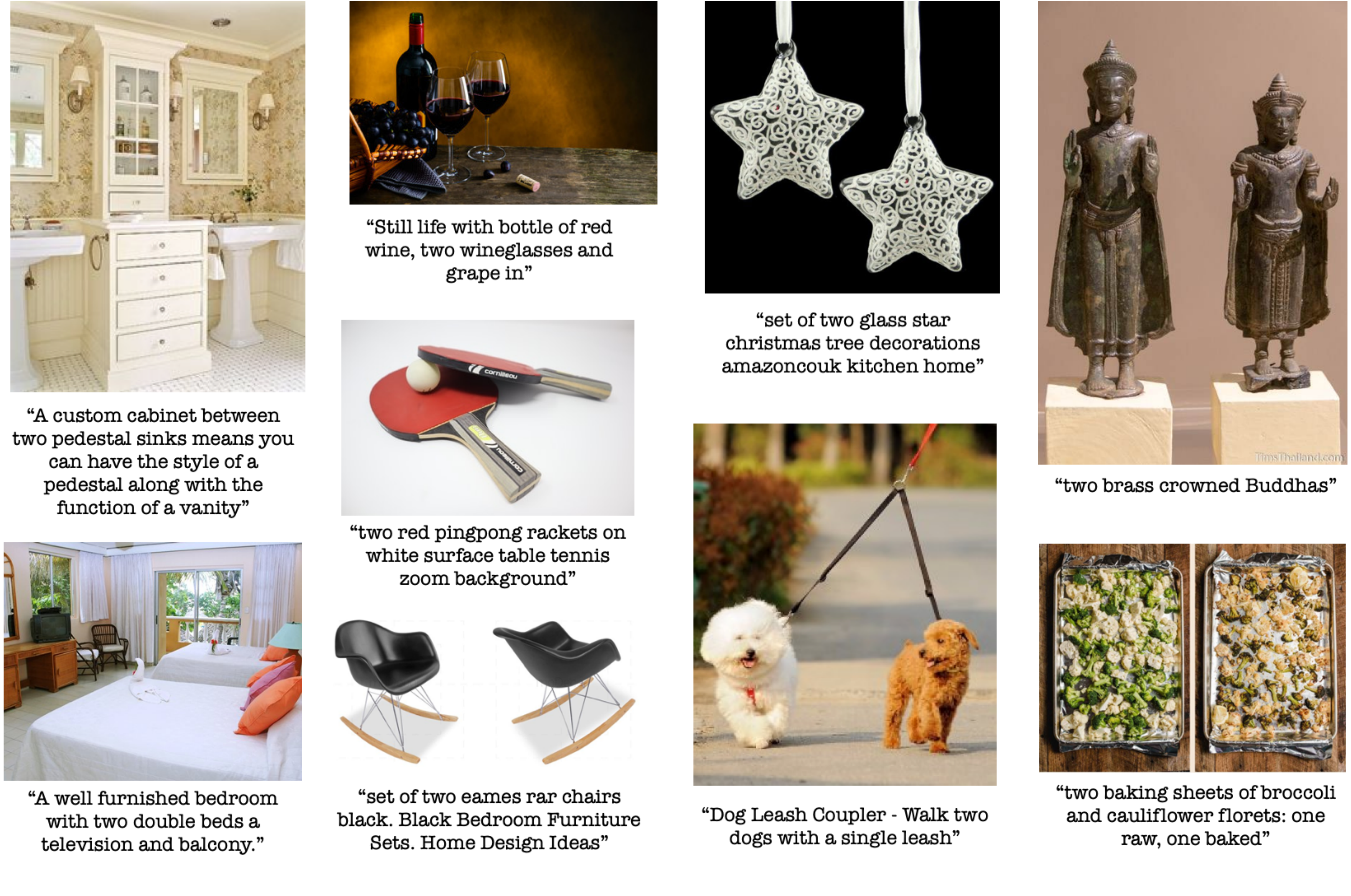}
  \vspace*{-0.2cm}
    \caption{\bf{Sampled images from CountBench labeled as ``two''.}}
    \label{fig:countbench_two}
  \vspace*{1cm}
\end{figure*}

\begin{figure*}[t!]
    \centering
    \includegraphics[width=0.90\linewidth]{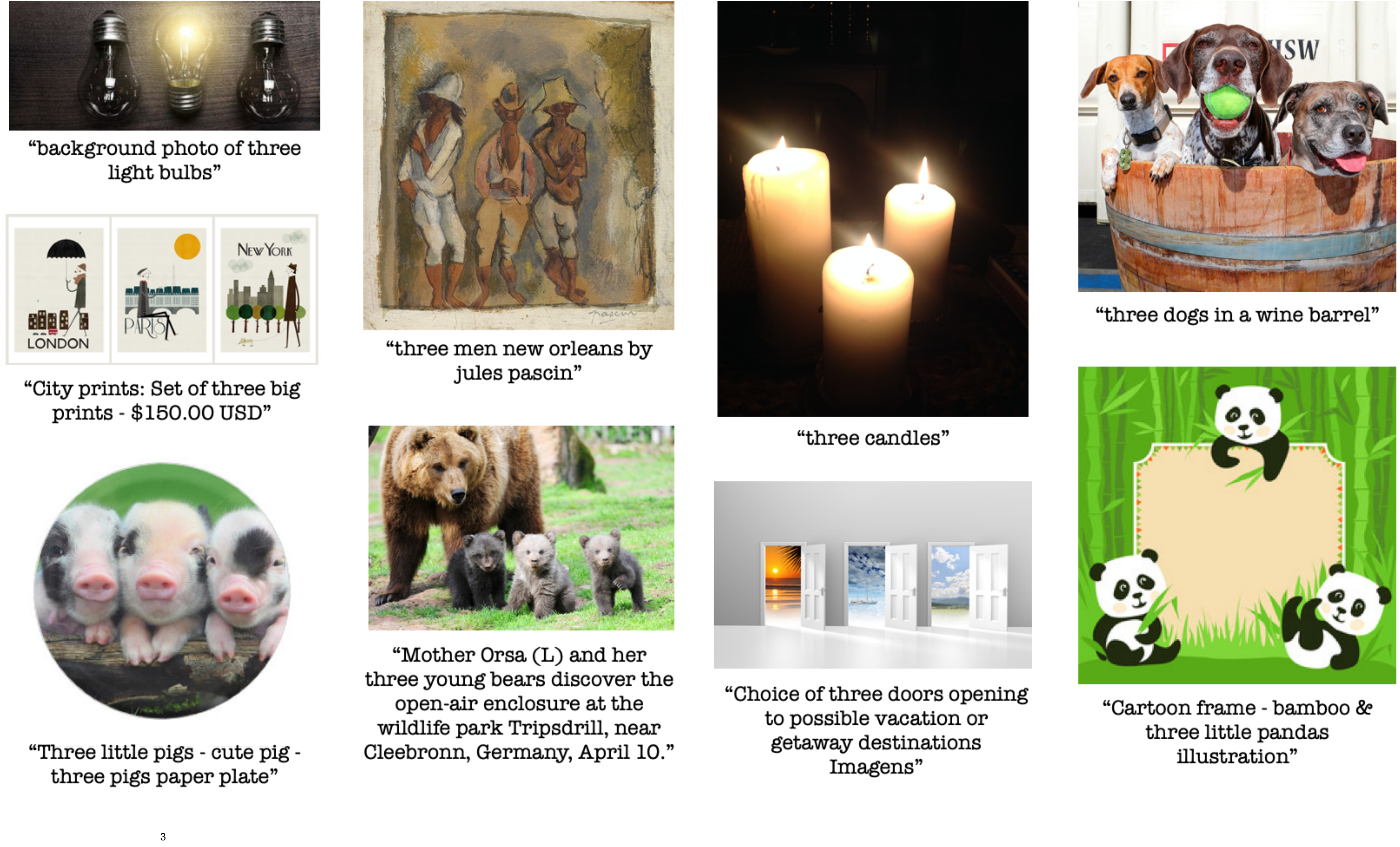}
  \vspace*{-0.2cm}
    \caption{\bf{Sampled images from CountBench labeled as ``three''.}}
    \label{fig:countbench_three}
  \vspace*{-0.2cm}
\end{figure*}

\begin{figure*}[t!]
    \centering
    \includegraphics[width=0.90\linewidth]{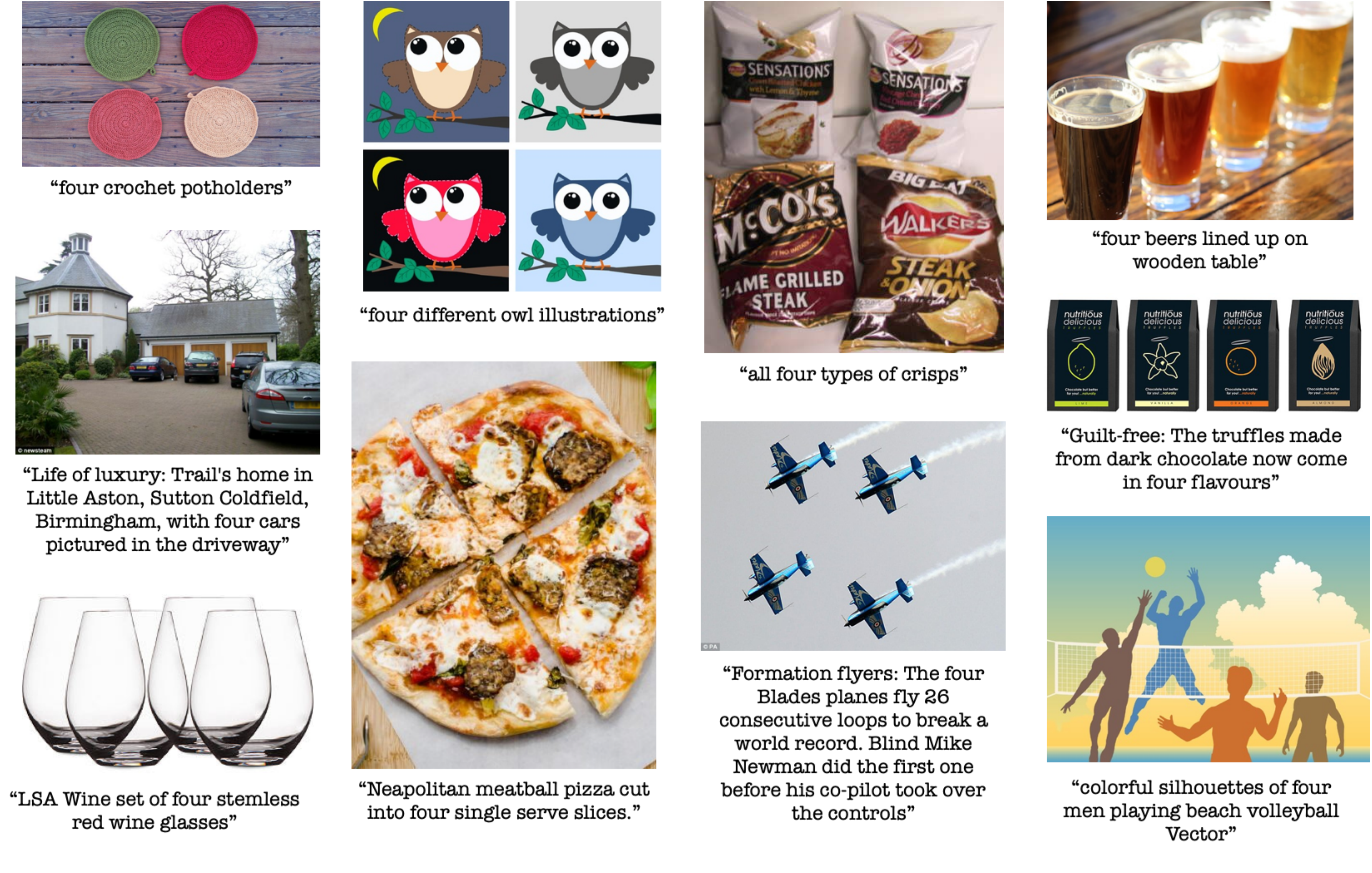}
  \vspace*{-0.2cm}
    \caption{\bf{Sampled images from CountBench labeled as ``four''.}}
    \label{fig:countbench_four}
  \vspace*{1cm}
\end{figure*}

\begin{figure*}[t!]
    \centering
    \includegraphics[width=0.90\linewidth]{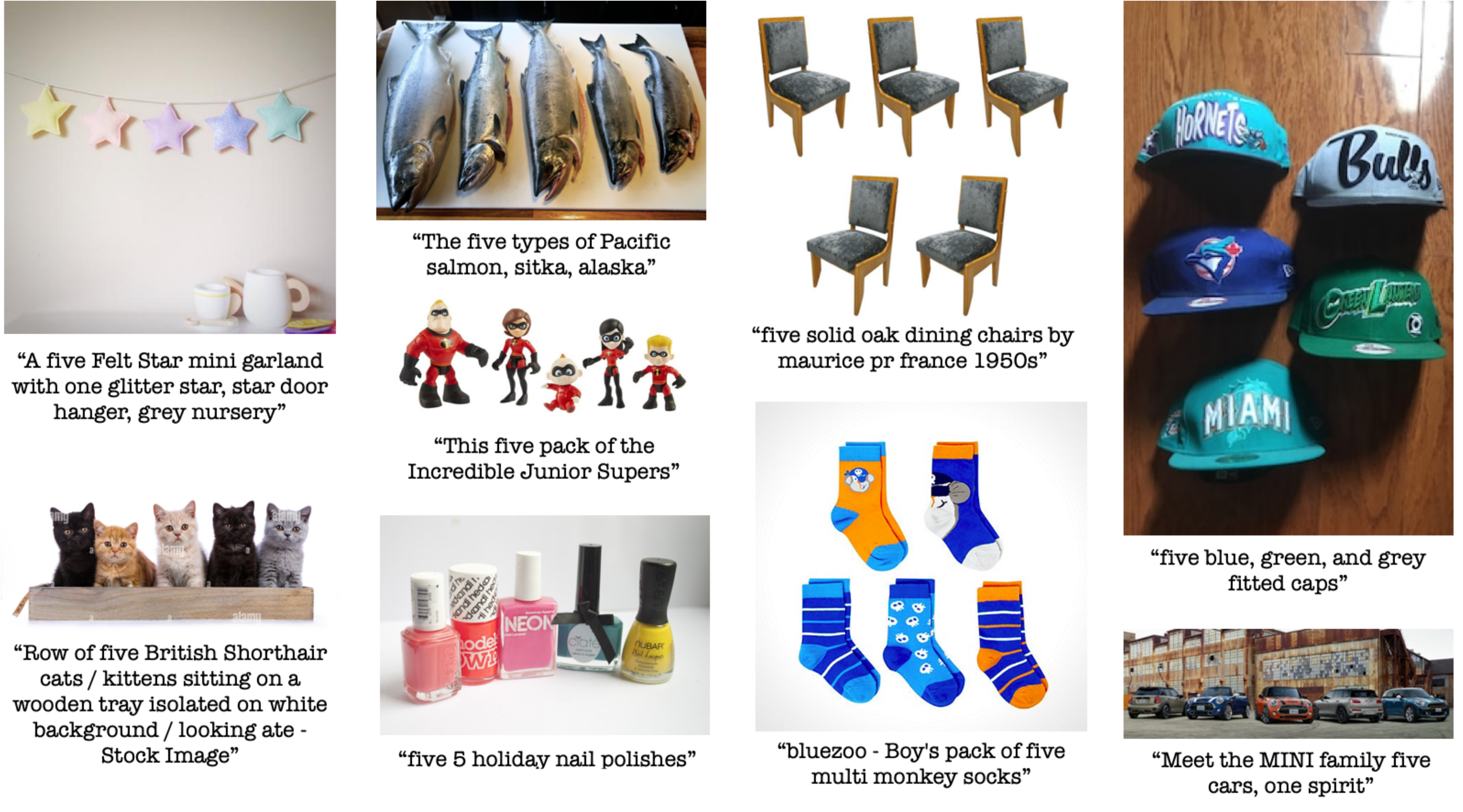}
  \vspace*{-0.2cm}
    \caption{\bf{Sampled images from CountBench labeled as ``five''.}}
    \label{fig:countbench_five}
    \vspace*{1cm}
\end{figure*}

\begin{figure*}[t!]
    \centering
    \includegraphics[width=0.90\linewidth]{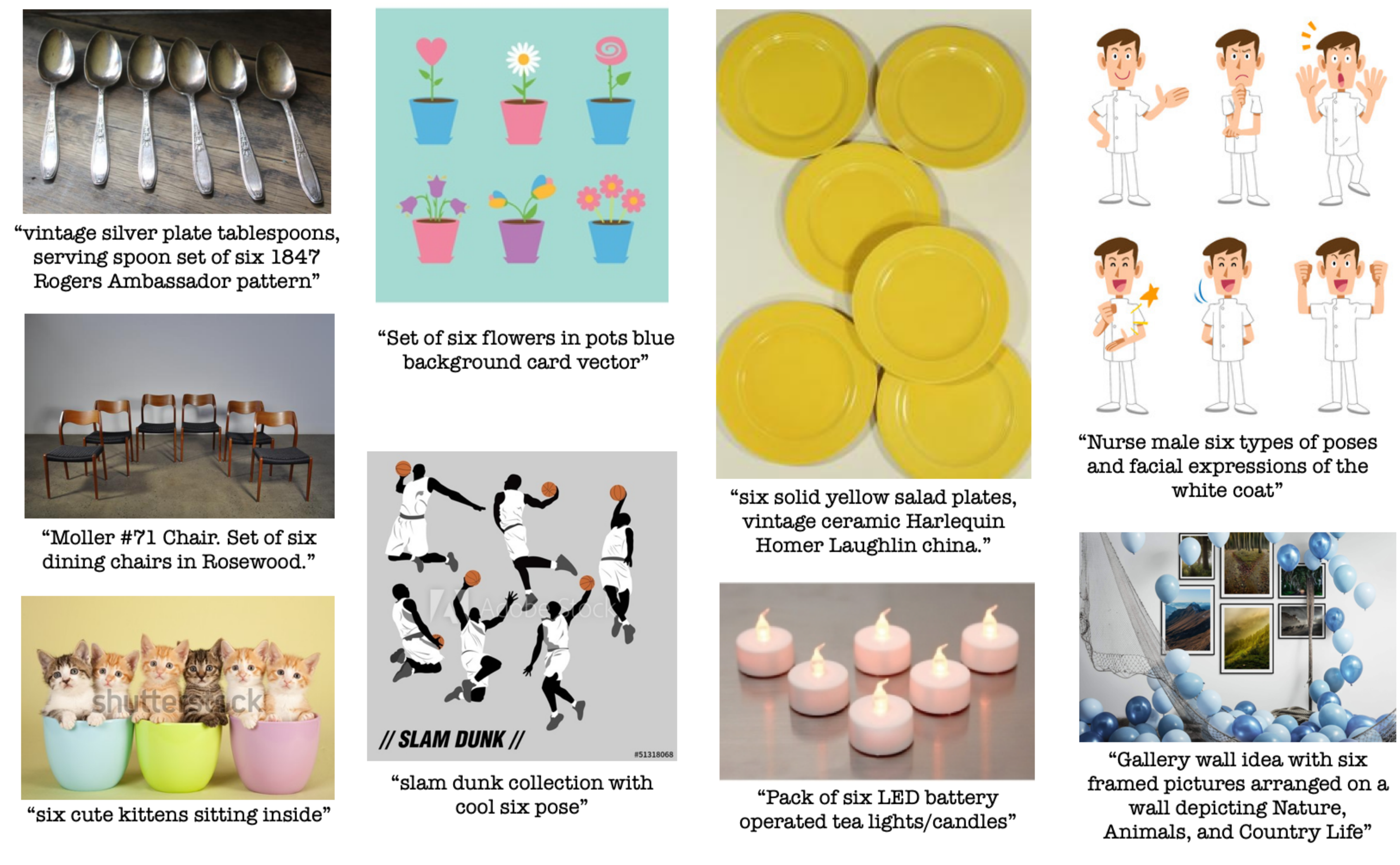}
  \vspace*{-0.2cm}
    \caption{\bf{Sampled images from CountBench labeled as ``six''.}}
    \label{fig:countbench_six}
\vspace*{0.5cm}
\end{figure*}

\begin{figure*}[t!]
    \centering
    \includegraphics[width=0.90\linewidth]{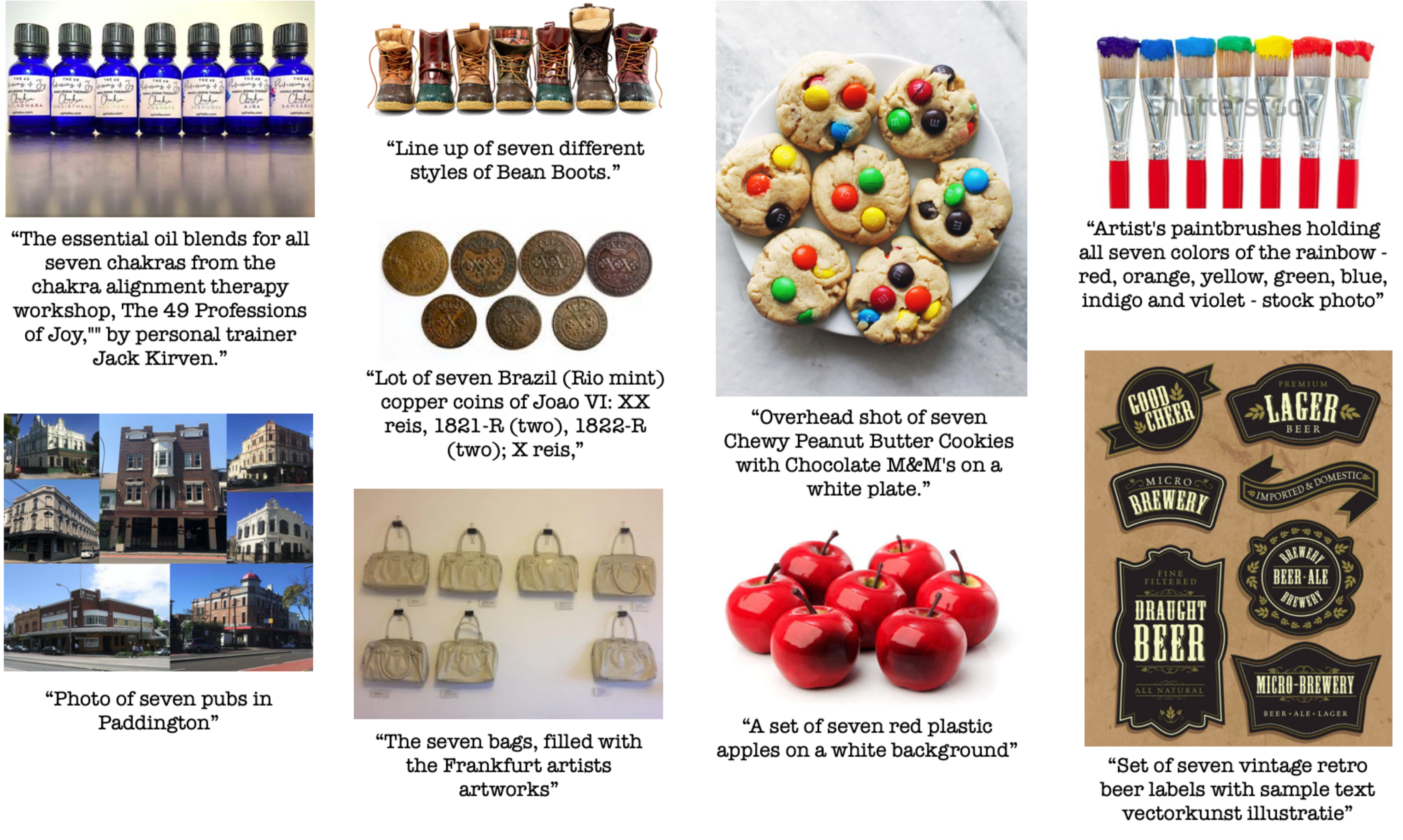}
  \vspace*{-0.2cm}
    \caption{\bf{Sampled images from CountBench labeled as ``seven''.}}
    \label{fig:countbench_seven}
    \vspace*{1cm}
\end{figure*}

\begin{figure*}[t!]
    \centering
    \includegraphics[width=0.90\linewidth]{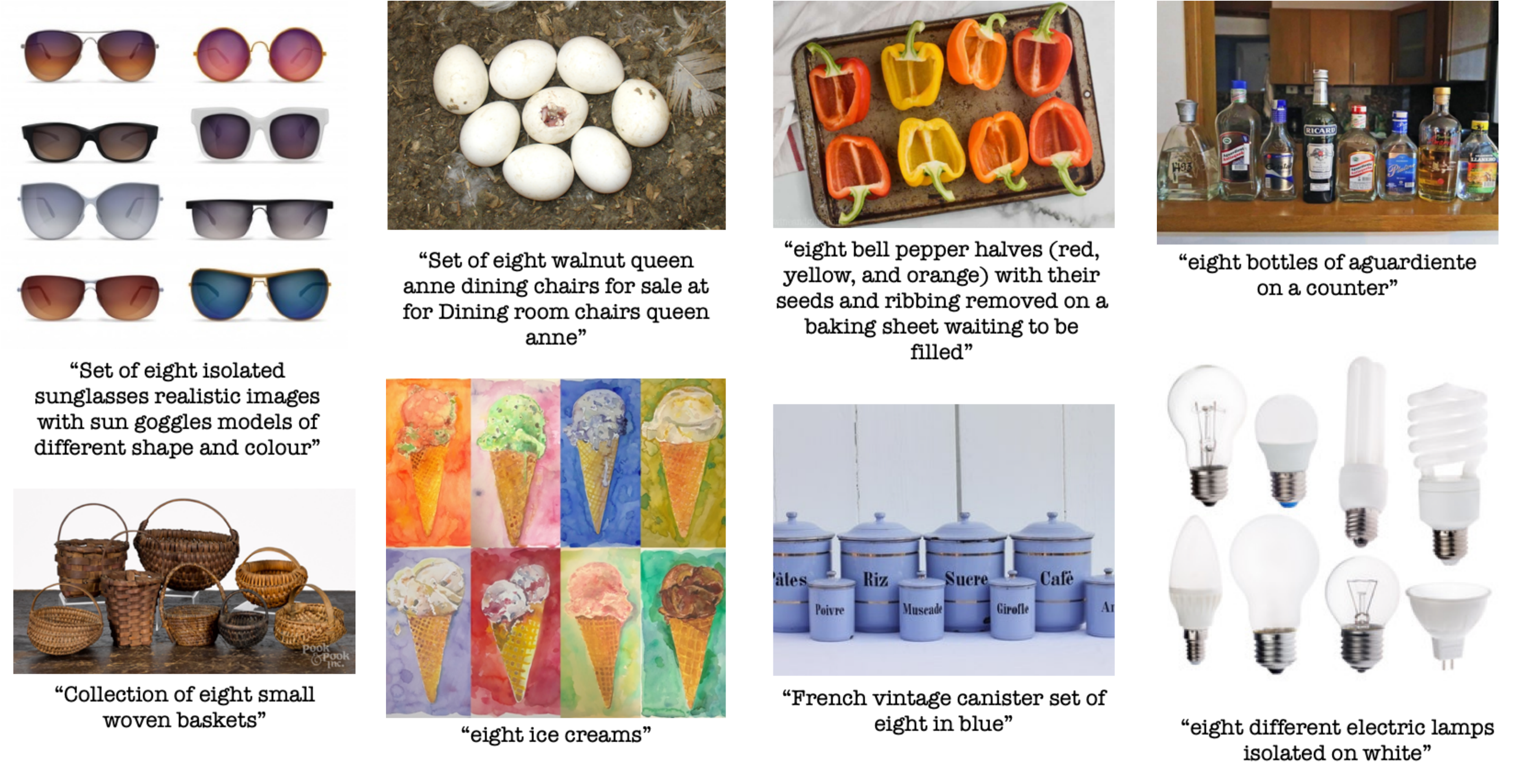}
  \vspace*{-0.2cm}
    \caption{\bf{Sampled images from CountBench labeled as ``eight''.}}
    \label{fig:countbench_eight}
    \vspace*{1cm}
\end{figure*}

\begin{figure*}[t!]
    \centering
    \includegraphics[width=0.90\linewidth]{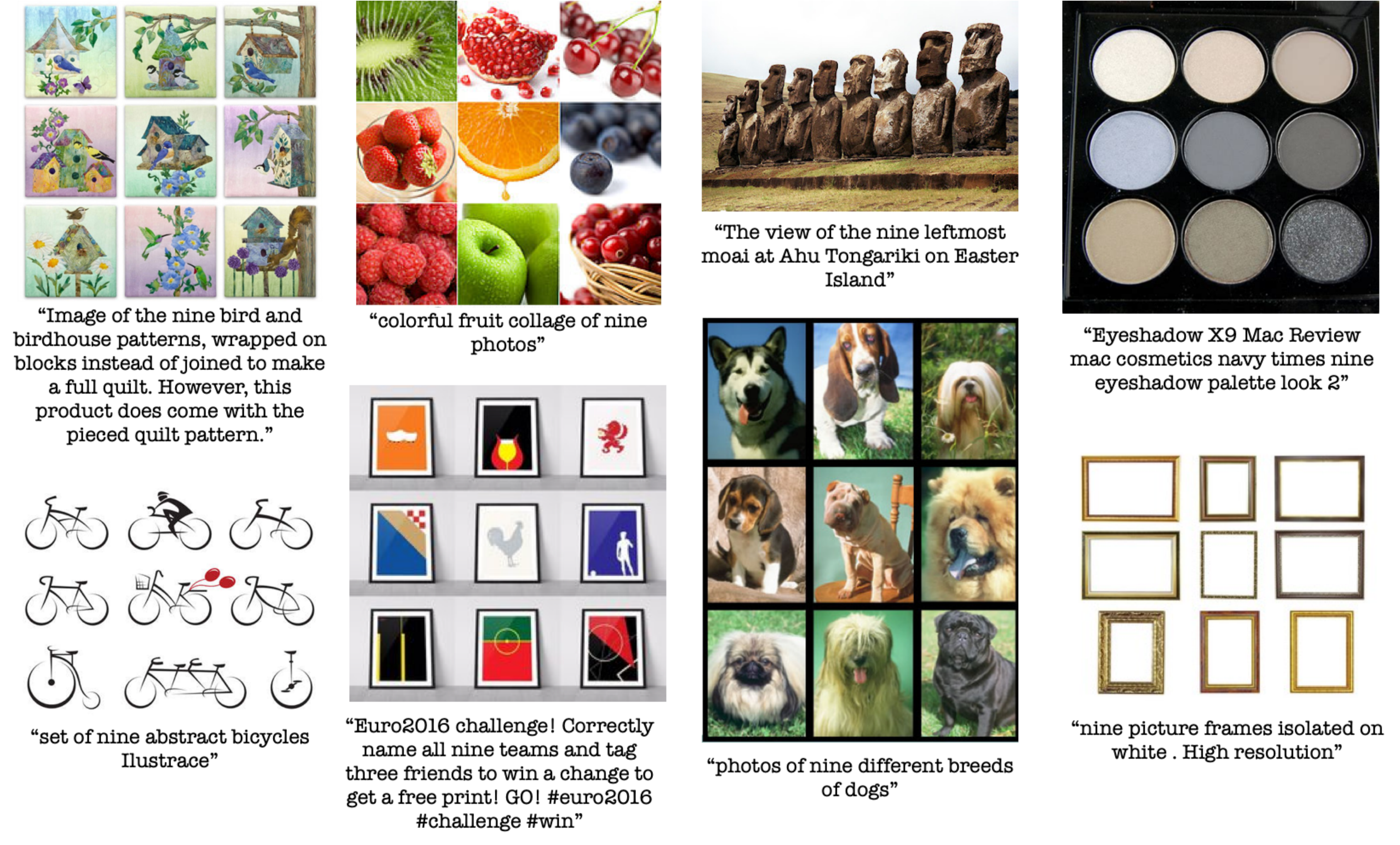}
  \vspace*{-0.2cm}
    \caption{\bf{Sampled images from CountBench labeled as ``nine''.}}
    \label{fig:countbench_nine}
    \vspace*{1.5cm}
\end{figure*}

\begin{figure*}[t!]
    \centering
    \includegraphics[width=0.90\linewidth]{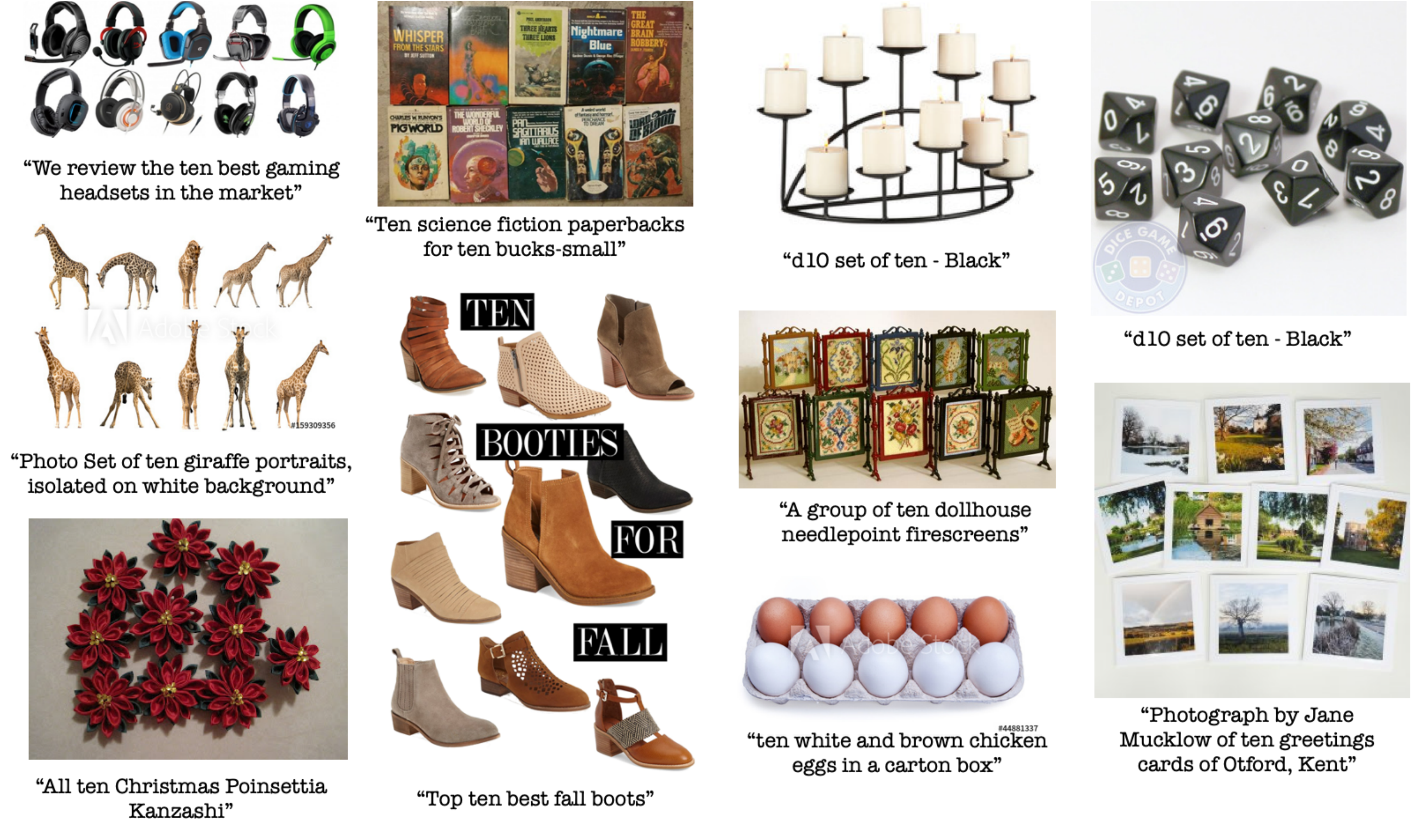}
  \vspace*{-0.2cm}
    \caption{\bf{Sampled images from CountBench labeled as ``ten''.}}
    \label{fig:countbench_ten}
\vspace*{1.5cm}
\end{figure*}

We include additional samples from {\it CountBench}, our automatically curated and manually verified object counting benchmark, which we plan to release. \cref{fig:countbench_two,fig:countbench_three,fig:countbench_four,fig:countbench_five,fig:countbench_six,fig:countbench_seven,fig:countbench_eight,fig:countbench_nine,fig:countbench_ten} showcase these additional image-caption pairs. The images vary in resolution and aspect ratios, and the captions vary in length.
Images labeled with larger numbers tend to be more grid-like (especially $``nine"$-labeled images as can be seen in \cref{fig:countbench_nine}). We believe that this can be attributed to the following: When a natural image contains a large number of objects ($>5-6$), it is more difficult to count them, and therefore the caption rarely contains a number. On the other hand, synthetically-created images with larger numbers of objects are usually created in a grid-like pattern, which facilitates much easier counting, leading to corresponding captions which often do contain the object count.

\label{app:countbench}

\end{document}